\documentclass[final,5p,times,twocolumn]{elsarticle}

\usepackage[utf8]{inputenc}
\usepackage[T1]{fontenc}
\usepackage{graphicx}
\usepackage{amsmath,amssymb,amsfonts}
\usepackage{algorithmic}
\usepackage{textcomp}
\usepackage{tabularx}
\usepackage{threeparttable}
\usepackage{xcolor}
\usepackage{array}
\usepackage{longtable}
\usepackage{booktabs}
\usepackage{adjustbox}
\usepackage{multirow}
\usepackage{siunitx}
\usepackage{tikz}
\usepackage{float}
\usepackage{subcaption}
\usepackage{url}
\usepackage[hidelinks]{hyperref}

\journal{Artificial Intelligence in Medicine}

\begin{document}
\begin{frontmatter}

\title{Anatomically Normalized Macular OCT Thickness Maps for Neurodegenerative Disease Classification: A Model Benchmark}
\author[i3a,diis]{Iker Morán-Cavero}
% \ead{imoran@unizar.es}

\author[i3a,diis]{Mónica Hernández}
% \ead{mhg@unizar.es}

\author[i3a,diis]{Elvira Mayordomo}
% \ead{elvira@unizar.es}

\author[hms,iisaragon]{Naiara Artiaga}
% \ead{email@ejemplo.es}

\author[hms,iisaragon]{Beatriz Cordón}
% \ead{email@ejemplo.es}

\author[hms,iisaragon]{Elena Garcia-Martin}
% \ead{email@ejemplo.es}

\affiliation[i3a]{
  organization={Aragon Institute of Engineering Research (I3A), University of Zaragoza},
  city={Zaragoza},
  country={Spain}
}

\affiliation[diis]{
  organization={Computer Science and Systems Engineering Department, University of Zaragoza},
  city={Zaragoza},
  country={Spain}
}

\affiliation[hms]{
  organization={Department of Ophthalmology, Miguel Servet University Hospital},
  city={Zaragoza},
  country={Spain}
}

\affiliation[iisaragon]{
  organization={Aragon Health Research Institute (IIS)},
  city={Zaragoza},
  country={Spain}
}

\begin{abstract}
Retinal optical coherence tomography (OCT) offers a non-invasive window into the central nervous system providing high-resolution cross-sectional images of the retina close to the optic nerve. In this study, we propose a comprehensive benchmark evaluating 2D high-resolution en-face macular thickness maps of the retinal nerve fiber layer (RNFL) and ganglion cell-inner plexiform complex (GCL-IPL) for the classification of neurodegenerative diseases (Multiple Sclerosis (MS), Parkinson's Disease (PD), and Alzheimer's Disease (AD)) from healthy controls (HC). By using an anatomically normalized cohort of 336 subjects acquired under the Fast Macula protocol, we evaluate multiple learning strategies over those en-face maps: traditional machine learning models (XGBoost), convolutional neural networks, variational autoencoders (VAEs) with perceptual loss, and self-supervised Vision Transformers under frozen feature extraction or LoRA fine-tuning strategies. Model evaluation was conducted using a strict nested 5-fold stratified group cross-validation with 10 macro-runs to eliminate data leakage and ensure statistical robustness. Additionally, we complemented the statistical evaluation with Monte Carlo Dropout uncertainty estimation and a qualitative evaluation with two interpretability frameworks (SHAP and Grad-CAM++). Our results demonstrate that GCL-IPL enfaces carry the primary discriminative signal, whereas RNFL alone suffers severe performance degradation in PD and AD. Across all disease cohorts, competitive performance was achieved, reaching peak AUCs of $0.82$ for MS, $0.73$ for PD, and $0.77$ for AD while maintaining low spatial uncertainty ($\mu_{\sigma} \leq 9\%$) under imbalanced scenarios.
\end{abstract}


\end{frontmatter}

\section{Introduction}
\label{sec:introduction}

Optical coherence tomography (OCT) has become an essential, non-invasive tool in current clinical practice, allowing for an in vivo, high-resolution visualization of the structure of the retina. Beyond its intrinsic value in ophthalmological evaluation, the retina is considered an accessible extension of the central nervous system. Consequently, structural alterations and thickness changes in its inner layers (such as the retinal nerve fiber layer (RNFL) or the ganglion cell layer-inner plexiform layer complex (GCL-IPL)) serve as critical biomarkers for the diagnosis and monitoring of different neurodegenerative pathologies. Among them, multiple sclerosis (MS), Parkinson's disease (PD), and Alzheimer's disease (AD) stand out for their clinical impact.

%  Resumen: critica a current metodos y desafios actuales con en-faces
Despite the volumetric structure of OCT images, standard clinical practice often relies on aggregated metrics, such as mean thicknesses of the retinal sectors (Nasal, Temporal, Superior, and Inferior). Although thickness maps are also available, their limited spatial resolution may obscure local changes and can be lost through the standard metrics. In contrast, high-resolution en-face thickness maps preserve more detailed spatial information, which may be useful for identifying disease-specific retinal alterations.

To address this limitation, our previous work introduced an anatomical normalization protocol integrated with a state-of-the-art retinal layer segmentation method~\cite{morancavero2026spatialnormalizationcrossdomainretinal}. By correcting acquisition tilts through an anatomical reference alignment (e.g., the foveal pit) and removing non-retinal structures such as the vitreous and choroid, this protocol establishes a consistent anatomical reference across the B-scans of each OCT volume. This alignment facilitates accurate retinal layer segmentation and enables the generation of high-resolution en-face thickness maps by projecting layer thickness measurements, computed from the segmented boundaries of individual B-scans, onto a two-dimensional representation. Unlike regional thickness measurements, these maps preserve local spatial patterns that may be lost through spatial averaging.

In this work, we evaluate whether these spatial patterns can help distinguish healthy controls from patients with neurodegenerative diseases. We compare different learning approaches, including traditional machine learning models, convolutional neural networks (CNNs), variational autoencoders (VAEs), and self-supervised Vision Transformers. These models are evaluated using small and imbalanced datasets.

The main contributions of this work are summarized in the following points:

\begin{itemize}
    \item An evaluation framework based on anatomically normalized high-resolution OCT en-face thickness maps for neurodegenerative diseases classification across Multiple Sclerosis (MS), Parkinson's Disease (PD), and Alzheimer's Disease (AD) cohorts.
    \item Benchmarking of multiple architectures: We compare conventional machine learning models, convolutional neural networks, VAE-based approaches, and self-supervised foundation models using high-resolution retinal thickness maps. We additionally evaluate both frozen and parameter-efficiently fine-tuned (LoRA) foundation backbones, allowing us to see the impact of parameter-efficient adaptation strategies in data-constrained clinical settings.
    \item Interpretability: We incorporate visual explainability techniques (SHAP and Grad-CAM++), allowing visualization of the retinal regions that contribute most to model predictions
    \item Layer-specific biomarker analysis: We investigate the discriminative value of the two most relevant retinal layers (RNFL and GCL-IPL) for neurodegenerative disease classification.
\end{itemize}

The remainder of this article is organized as follows: Section \ref{sec:related_work} reviews the literature and related work. Section \ref{sec:materials_methods} details the OCT acquisition protocol and datasets used and the classification architectures explored. Section \ref{sec:results} presents the quantitative results and statistical significance analysis along with a explainability and generalization consistency subsection. Section \ref{sec:discussion} discusses the findings in the current clinical context along with the limitations and future directions of this work. Finally, Section \ref{sec:conclusion} concludes the article with a summary of the main findings.

\section{Related work}
\label{sec:related_work}

% Nota: Revisar este capítulo
\subsection{Retina involvement on neurodegenerative diseases, biomarkers, and clinical relevance}

%  biomarcadores y justificación de su uso en neurodegeneración -  meto algo de Steinert aunque no estemos trabajando con ellas en este documento??
Optical Coherence Tomography (OCT) has traditionally focused on the study of the retina as a window to the eye, but, in recent years, the field of oculomics\footnote{Medical field that uses eye, in particular, the retina as a window to diagnose neurodegenerative diseases} has emerged, considering the retina as an accessible extension of the central nervous system and exploiting retinal imaging information to identify biomarkers of neurological diseases. In this context, optical coherence tomography (OCT) has demonstrated that thinning of the inner retinal layers (in particular, the ones having neurons) constitutes a reproducible structural biomarker of different neurodegenerative diseases. In multiple sclerosis (MS), studies and analyses have confirmed a significant reduction in the thickness of the peripapillary retinal nerve fiber layer (pRNFL), macular RNFL, and ganglion cell layer-inner plexiform layer complex (GCL-IPL), being this effect especially pronounced in eyes with a history of optic neuritis \cite{petzold2017retinal,bsteh2023diagnostic}. Regarding Parkinson's disease (PD), several studies have described a progressive yet statistically significant thinning of both the RNFL and the GCL-IPL, which is further associated with clinical progression, cognitive decline, and motor impairment \cite{satue2014retinal,chrysou2019retinal, garcia2014distribution}. Although these morphological changes are often more diffuse and subtle than the focal lesions observed in MS, they represent crucial structural biomarkers that benefit significantly from advanced high-resolution spatial analysis. Likewise, consistent reductions in retinal thickness have been reported in Alzheimer's disease (AD). The landmark meta-analysis by Thomson et al.~\cite{thomson2015systematic} identified a significant thinning of the RNFL compared to healthy controls. Furthermore, subsequent large-scale meta-analyses and multi-cohort prospective studies have reported a pronounced reduction in the thickness of the GCL-IPL complex, robustly associating this inner retinal impairment with cognitive decline, hippocampal atrophy, and overall disease progression~\cite{chen2023association, ashraf2023retinal, van2024associations}.

\subsection{Deep learning and convolutional neural networks using retinal OCT data for neurodegenerative disease analysis}

Automatic disease detection in retinal optical coherence tomography (OCT) follows two main computational approaches. The first relies on hand-crafted measurements (such as the mean thickness of the retinal layers) extracted from the clinical software and fed as tabular data into traditional machine learning classifiers~\cite{hasanshahi2026ensemble}. While this approach is computationally efficient, it loses the spatial context and subtle details present in the original scans. The second approach utilizes raw OCT data (either 2D B-scans or complete 3D volumes) using convolutional neural networks (CNNs). Although raw data architectures preserve spatial structure, they are difficult to optimize. Fully 3D CNNs require massive annotated datasets and heavy computational power. In data-constrained clinical settings, 2D image-based deep learning have shown promising performance for detecting neurodegenerative diseases from retinal imaging data~\cite{cheung2022deep}.

Furthermore, raw 3D OCT volumes are highly susceptible to geometric tilt during scan acquisition, introducing subtle spatial distortions that can degrade learned accuracy. Addressing these challenges in neurodegenerative disease characterization, recent frameworks have shifted toward image-based deep learning operating on extracted structural layers. For instance, in multiple sclerosis (MS), Khodabandeh et. al~\cite{khodabandeh2026deep} demonstrated that while raw cross-sectional B-scans exhibit subtle alterations often indistinguishable from healthy controls, training deep networks on 2D combined retinal thickness maps yields superior discriminatory power and highly accurate diagnostic classification. Similarly, in Parkinson's disease (PD) and related neurodegenerative disorders, end-to-end convolutional architectures trained directly on 2D macular structural maps and multimodal retinal views~\cite{richardson2024multimodal} have proven highly effective in capturing complex, distributed spatial patterns of neuroretinal thinning linked to clinical progression.

Recent studies have further explored the diagnostic potential of retinal structural information extracted from OCT volumes. Álvarez-Rodríguez et al.~\cite{ALVAREZRODRIGUEZ2024103006} proposed a fully automatic deep learning framework for screening multiple neurodegenerative diseases using segmented retinal layers from macular and optic disc OCT scans. Their findings highlighted the benefits of combining complementary retinal views and using the layer-specific structural information for disease classification.

\subsection{Foundational models}

Beyond standard convolutional architectures, retinal image analysis has progressively evolved toward self-supervised learning (SSL) and vision foundation models. Earlier studies in neurodegenerative diseases primarily relied on supervised deep learning applied to 2D retinal imaging. For example, in Alzheimer’s disease, deep learning models trained on retinal photographs achieved promising diagnostic performance, demonstrating the potential of non-invasive retinal biomarkers for disease screening~\cite{cheung2022deep}. More recent works have incorporated attention mechanisms, SSL strategies, and foundation-model based representations to capture richer spatial and semantic retinal features~\cite{kesu2025alzheimer,jamshidiha2025explainable}.

%  CRITICAR MAS A.C?? el zhou este es bueno realmente?? He añadido una comparativa pero meh, no se parece mucho al nuetro
Building on these principles, retinal foundation models and self-supervised learning (SSL) models have emerged to reduce the need for massive annotated clinical datasets. A prominent example is RETFound~\cite{zhou2023foundation}, which employed a masked autoencoder (generative self-supervision) to pre-train a Vision Transformer using approximately 1.6 million unlabelled retinal images, including both fundus photographs and optical coherence tomography (OCT) scans. It achieved state-of-the-art performance across multiple retinal imaging tasks, including the detection of ocular diseases such as glaucoma, as well as the prediction of neurological conditions, including Parkinson's disease. These advancements justify the deployment of self-supervised Vision Transformers and generative frameworks (such as Variational Autoencoders) to extract robust, disease-discriminative representations from high-resolution structural retinal maps~\cite{alvarez2024fully}.

Despite these advancements, there is still a lack of frameworks that evaluate and compare how these different feature extraction strategies perform a binary classification task for multiple neurodegenerative diseases using high-resolution en-face thickness maps taking into account the challenges of limited sample sizes, potential class imbalance and the lack of large-scale annotated datasets that lead to a possible overfitting of the models. In this work, we aim to address this gap by proposing an evaluation of different deep learning architectures and feature extraction strategies for the classification of multiple neurodegenerative diseases using high-resolution en-face retinal thickness maps obtained from optical coherence tomography (OCT) scans under the same experimental conditions by applying a consistent normalization protocol before the layer segmentation and, subsequently, the en-face thickness map generation.

\section{Materials and Methods}
\label{sec:materials_methods}

\subsection{Data Acquisition and Cohort Description}

\subsubsection{OCT Acquisition Protocols}

\begin{figure}[t]
    \centering
    \includegraphics[width=\linewidth]{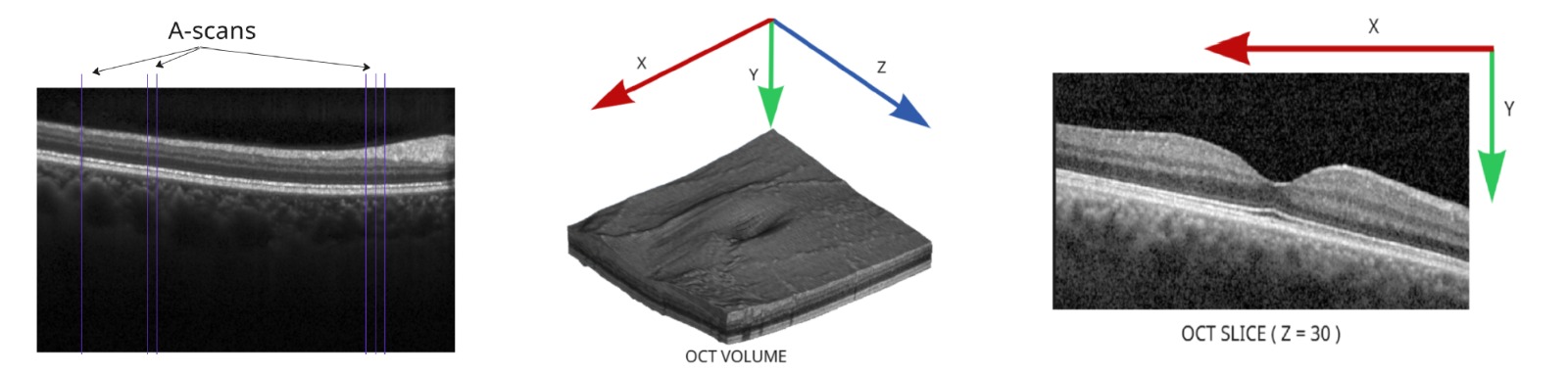}
    \caption{Illustration of OCT acquisition units. Left: an A-scan, a one-dimensional depth profile. Center: Volumetric OCT: a three-dimensional representation constructed from a series of B-scans across the region of interest. Right: a B-scan, a two-dimensional cross-sectional image formed by laterally sampling multiple A-scans. }
    \label{fig:types}
\end{figure}

\begin{figure}[t]
    \centering
    \includegraphics[width=0.5\linewidth, clip, trim=0 0 900 0]{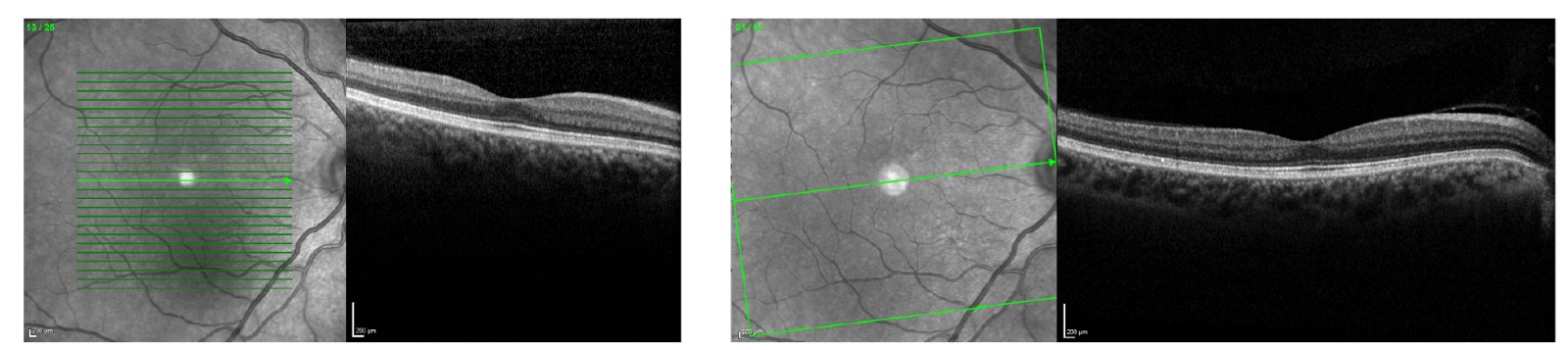}
    \caption{Example of the Fast Macula (FastMac) OCT acquisition protocol. Left, a fundus image with the scan pattern overlaid in green. Right, the corresponding B-scan from the current cut highlighted with an intense green on the fundus image.}
    \label{fig:prots}
\end{figure}

% NOTA: El protocolo ppole lo describiiría en la parte de futuro trabajo donde la generalización así como la patología Steinert
Optical Coherence Tomography (OCT) builds volumetric representations hierarchically: individual axial depth profiles (A-scans) form two-dimensional cross-sections (B-scans), which collectively form a 3D volume (Figure \ref{fig:types}). In this study, we utilize the Fast Macula (FastMac) acquisition protocol from the Heidelberg Spectralis system (Figure \ref{fig:prots}), which provides a rapid, fovea-centered acquisition tailored for routine clinical evaluation with minimal scan times.

\subsubsection{Retinal Layer Segmentation}

For the segmentation task, we used an annotated dataset from Johns Hopkins University (JHU) \cite{JHU}. The dataset consists of 35 OCT volumes acquired with the FastMac protocol, each containing 49 B-scans. The cohort includes 14 healthy controls and 21 multiple sclerosis patients, with manual delineations of nine retinal boundaries provided for each volume. The segmentation models, presented in \cite{morancavero2026spatialnormalizationcrossdomainretinal}, were trained on this dataset to learn the anatomical structure of the retina and to generate accurate thickness maps for subsequent classification tasks. Specifically, we used the SD-LayerNet model, which is a state-of-the-art architecture for retinal layer segmentation in OCT images \cite{sdlayernet}. This model was initally inspired in the SDNet, which proposed a segmentation strategy based on disentangled representations and incorporated an unsupervised branch to learn the representation of the input images. SD-LayerNet integrates these ideas with the incorporation of anatomical priors specific for the retinal layer segmentation such as layer positions, layer continuity and shape regularizers. These priors guide the supervised segmentation module while the layer predictions generated condition the autoencoder branch, making a joint learning strategy where both modules benefit from each other. 

\subsubsection{Classification Datasets}

\begin{table}[t]
\centering
\begin{threeparttable}
\footnotesize
\setlength{\tabcolsep}{6pt}
\caption{Dataset breakdown showing initial ($N_0$) and final ($N$) patient counts.}
\label{tab:dataset_breakdown}

\begin{tabular}{@{}lcc@{}}
\toprule
\textbf{Cohort} &
\shortstack{\textbf{Initial}\\\textbf{Patients} ($N_0$)} &
\shortstack{\textbf{Final}\\\textbf{Patients} ($N$)} \\
\midrule
Multiple Sclerosis (MS) & 152 & 149 \\
Parkinson's Disease (PD) & 82 & 79 \\
Alzheimer's Disease (AD) & 27 & 23 \\
\midrule
Healthy Controls (HC) & 75 & 59 \\
\midrule
\textbf{Total} & \textbf{336} & \textbf{310} \\
\bottomrule
\end{tabular}

\begin{tablenotes}[para,flushleft]
\footnotesize
\textit{Note:} Each patient contributes two eyes (OD and OI) from a single visit, resulting in 620 eyes.
\end{tablenotes}

\end{threeparttable}
\end{table}

To evaluate the classification performance we used an extensive multi-disease dataset collected over a 10-year period during routine clinical practice at Miguel Servet Hospital (Zaragoza, Spain). This cohort includes healthy controls (HC) alongside patients diagnosed with three neurodegenerative conditions: Multiple Sclerosis (MS), Parkinson's disease (PD), and Alzheimer's disease (AD). In addition, to ensure dataset integrity and prevent longitudinal data leakage, we implemented a single-visit, bilateral filtering strategy. Specifically, we selected the latest available visit containing valid scans for both eyes (OD and OI). If a patient presented data for only one eye during their most recent session, the selection algorithm automatically rolled back to their latest visit with complete bilateral acquisitions. Consequently, 100\% of the included patients contribute exactly two samples, eliminating single-eye scans and multi-visit duplicates.

All subjects underwent routine examinations and were selected based on strict inclusion criteria: best-corrected visual acuity $\ge$ 0.5 (Snellen), refractive errors $< \pm$5 dioptres of spherical equivalent, and no history of concomitant ocular pathologies (e.g., glaucoma). Diagnosis for MS, PD, and AD followed standard clinical, neuroimaging, and cognitive criteria (e.g., DSM-IV, Hoehn \& Yahr scale, and United Kingdom Brain Bank). Quality control filters were strictly applied, discarding scans below standard clinical quality thresholds ($< 25/40$). All procedures adhered to the Declaration of Helsinki and were approved by the local Ethics Committee (CEICA, ref: C.I. PI21/113). Table \ref{tab:dataset_breakdown} summarizes the final dataset composition after applying the aforementioned filtering criteria, resulting in a total of 310 unique patients (620 eyes) across all cohorts.

\subsection{Retinal Thickness Representation and Preprocessing}

\subsubsection{En-face generation}

\begin{figure}[t]
    \centering
    \includegraphics[width=\linewidth]{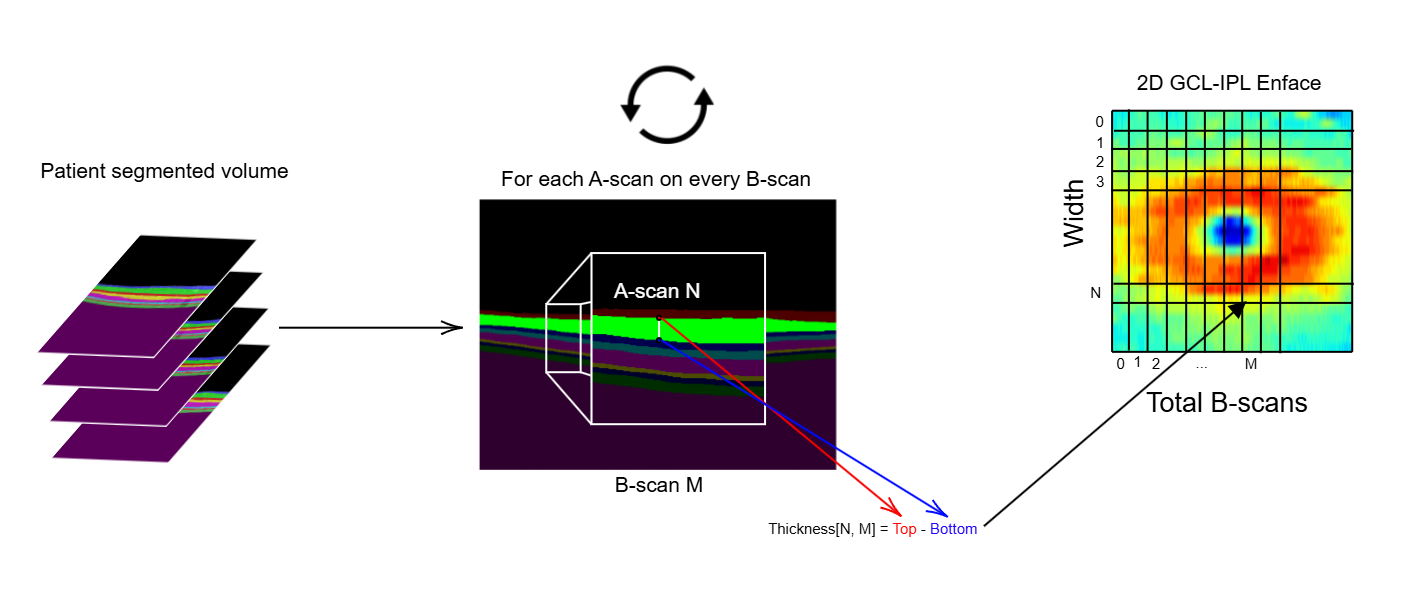}
    \caption{Pipeline for en-face generation. Starting from the segmentations obtained with SD-LayerNet, we calculate the thickness of the GCL-IPL layer at each A-scan as the distance between the top and bottom of the layer. This results in a 2D en-face map representing the thickness distribution across the retina.}
    \label{fig:enface}
\end{figure}

\begin{figure}[t]
    \centering
    \begin{minipage}{0.49\linewidth}
        \centering
        \includegraphics[
            width=\linewidth,
            trim=50 50 50 50,
            clip
        ]{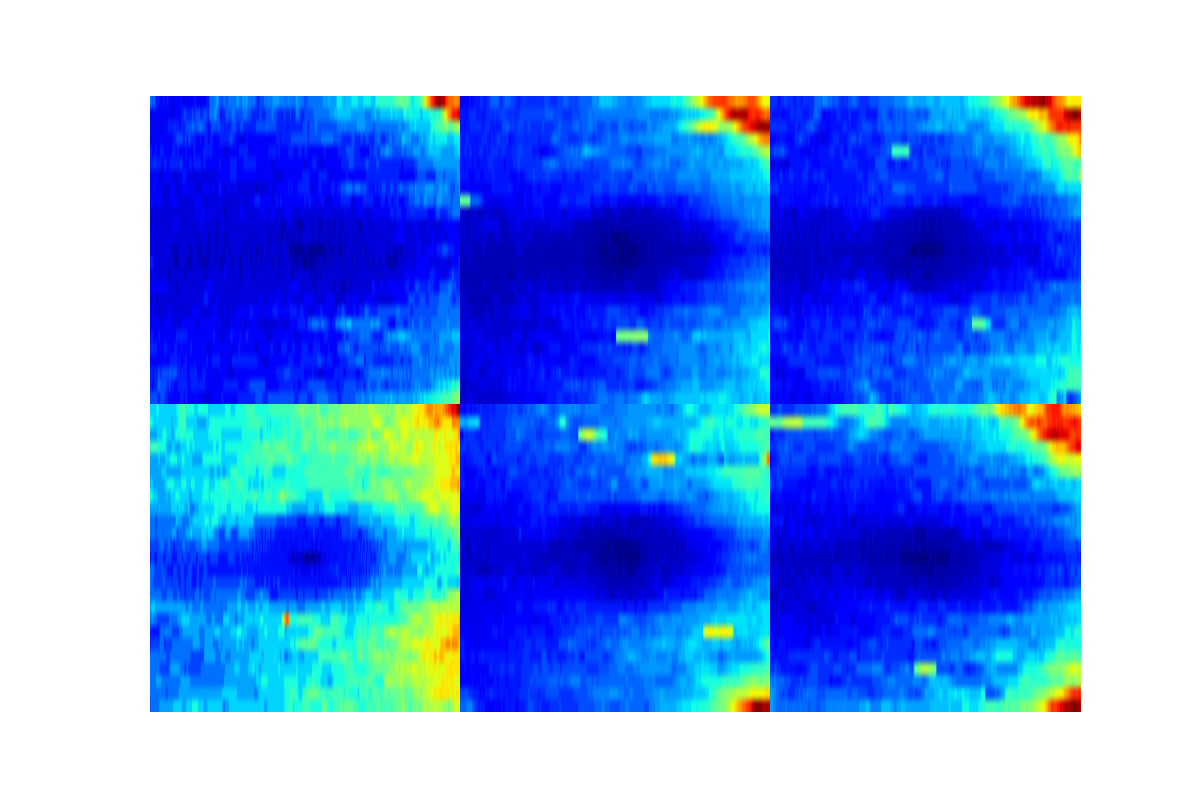}
    \end{minipage}
    \hfill
    \begin{minipage}{0.49\linewidth}
        \centering
        \includegraphics[
            width=\linewidth,
            trim=50 50 50 50,
            clip
        ]{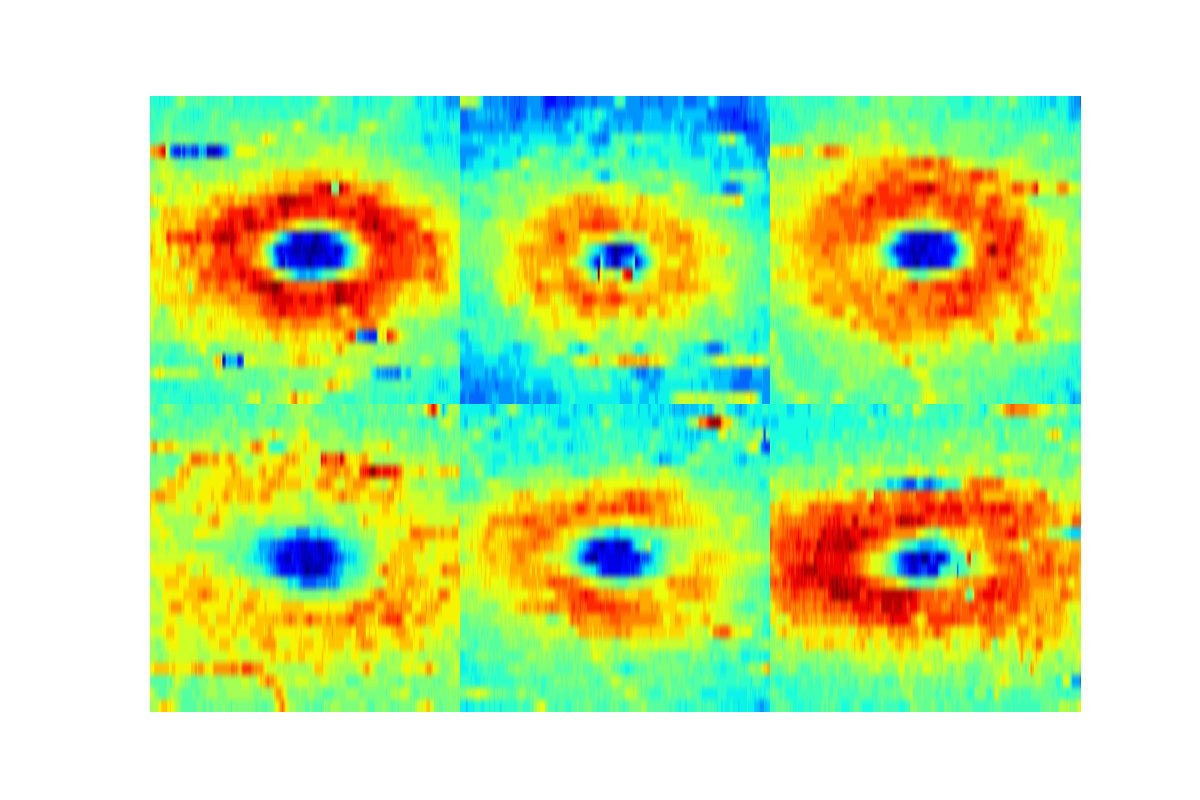}
    \end{minipage}
    \caption{En-face examples obtained from segmentations of MS patients. Left: RNFL en-face maps. Right: GCL-IPL en-face maps.}
    \label{fig:enface_example}
\end{figure}

Using the retinal layer segmentations obtained from SD-LayerNet~\cite{morancavero2026spatialnormalizationcrossdomainretinal}, we generated high-resolution 2D en-face thickness maps for each retinal layer. For each A-scan, layer thickness was computed as the vertical distance between its upper and lower segmented boundaries. These measurements were then projected into a 2D grid, preserving the spatial resolution of the original segmentations without regional averaging. The anatomical normalization incorporated before the retinal layer segmentation provides a consistent reference frame across the B-scans of each OCT volume, facilitating the generation of spatially coherent thickness maps. Figure~\ref{fig:enface} illustrates this process, while representative RNFL and GCL-IPL maps are shown in Figure~\ref{fig:enface_example}.

\subsubsection{Post-Processing and Layer Selection}

\begin{figure}[t]
    \centering
    \includegraphics[width=\linewidth]{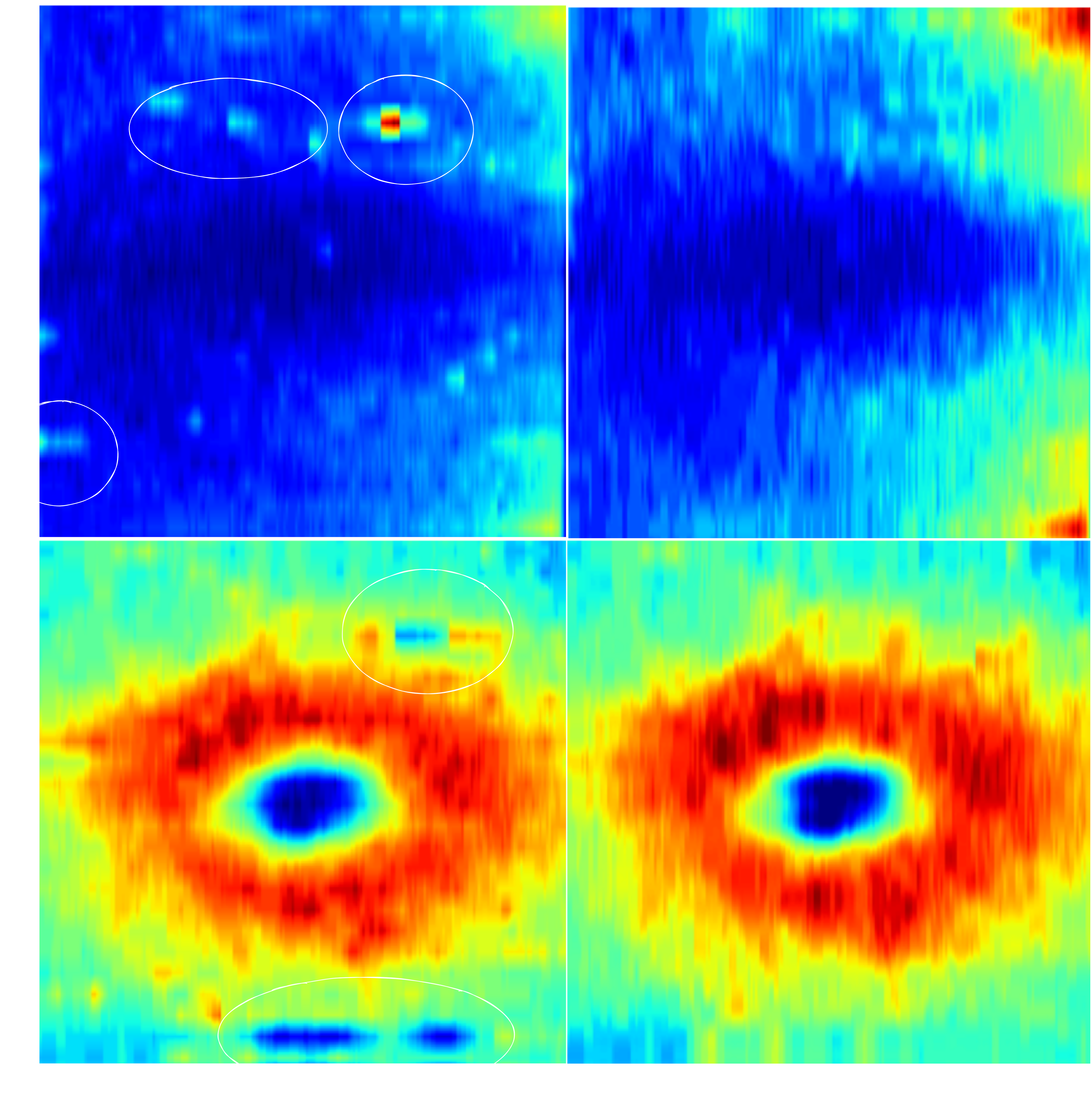}
    \caption{Effect of post-processing on en-faces. Lower Row: GCL-IPL en-faces. Upper row: RNFL en-faces. Left column: Original en-faces with segmentation artifacts highlighted with white circles. Right column: Post-processed en-faces after applying a median filter.}
    \label{fig:enface_post}
\end{figure}

Although spatial normalization mitigates most of the structural artifacts, localized inter-slice segmentation inconsistencies may be still present. To refine these thickness maps, we applied an asymmetric median filter with $3 \times 1$, kernel along the inter-slice direction to mitigate thickness discontinuities arising from slice-wise segmentation errors (Figure~\ref{fig:enface_post}).

\begin{figure}[t]
    \centering
    \begin{subfigure}[t]{0.49\linewidth}
        \centering
        \includegraphics[width=\linewidth]{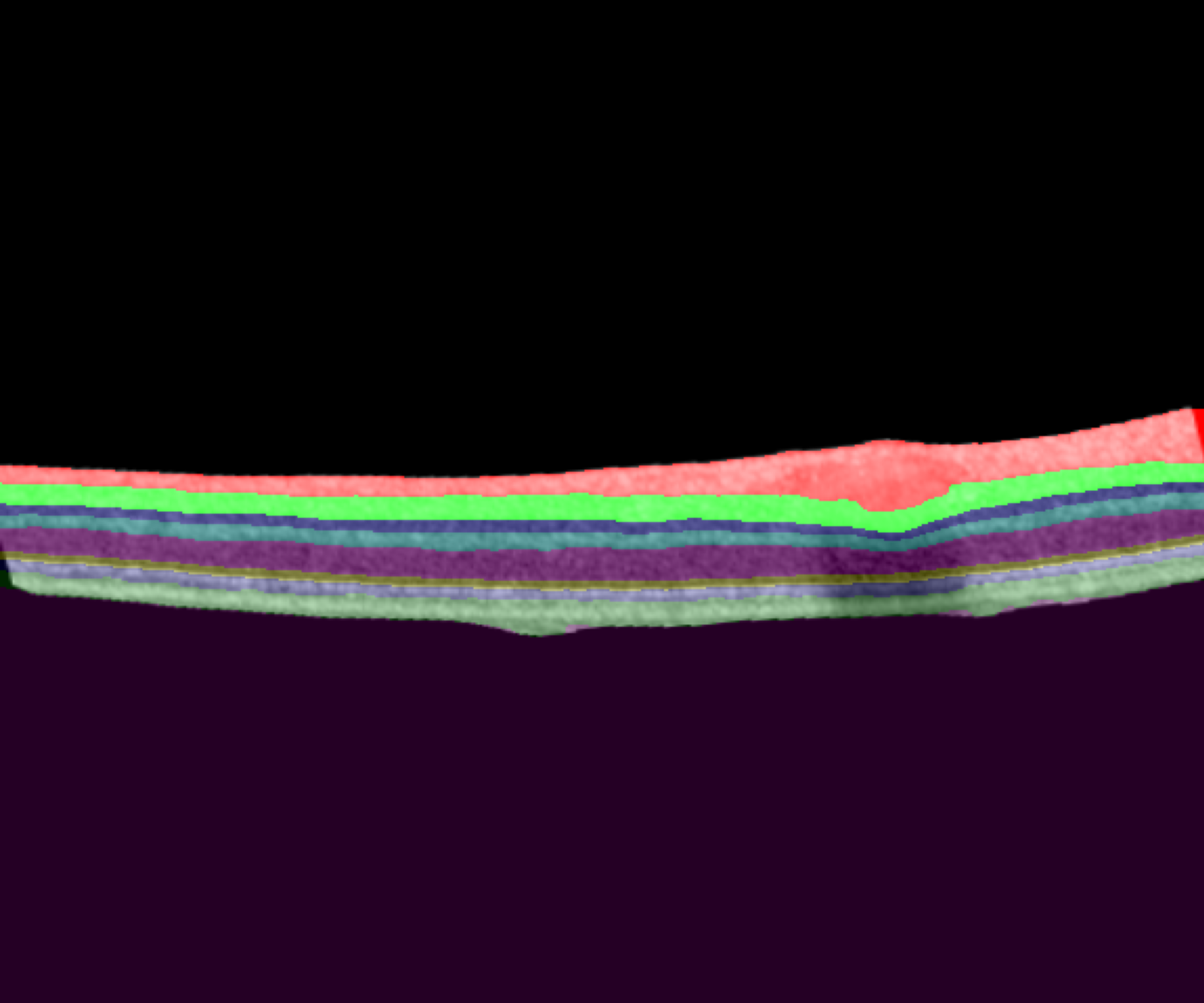}
        \caption{Non-foveal B-scan}
        \label{fig:seg1}
    \end{subfigure}
    \hfill
    \begin{subfigure}[t]{0.49\linewidth}
        \centering
        \includegraphics[width=\linewidth]{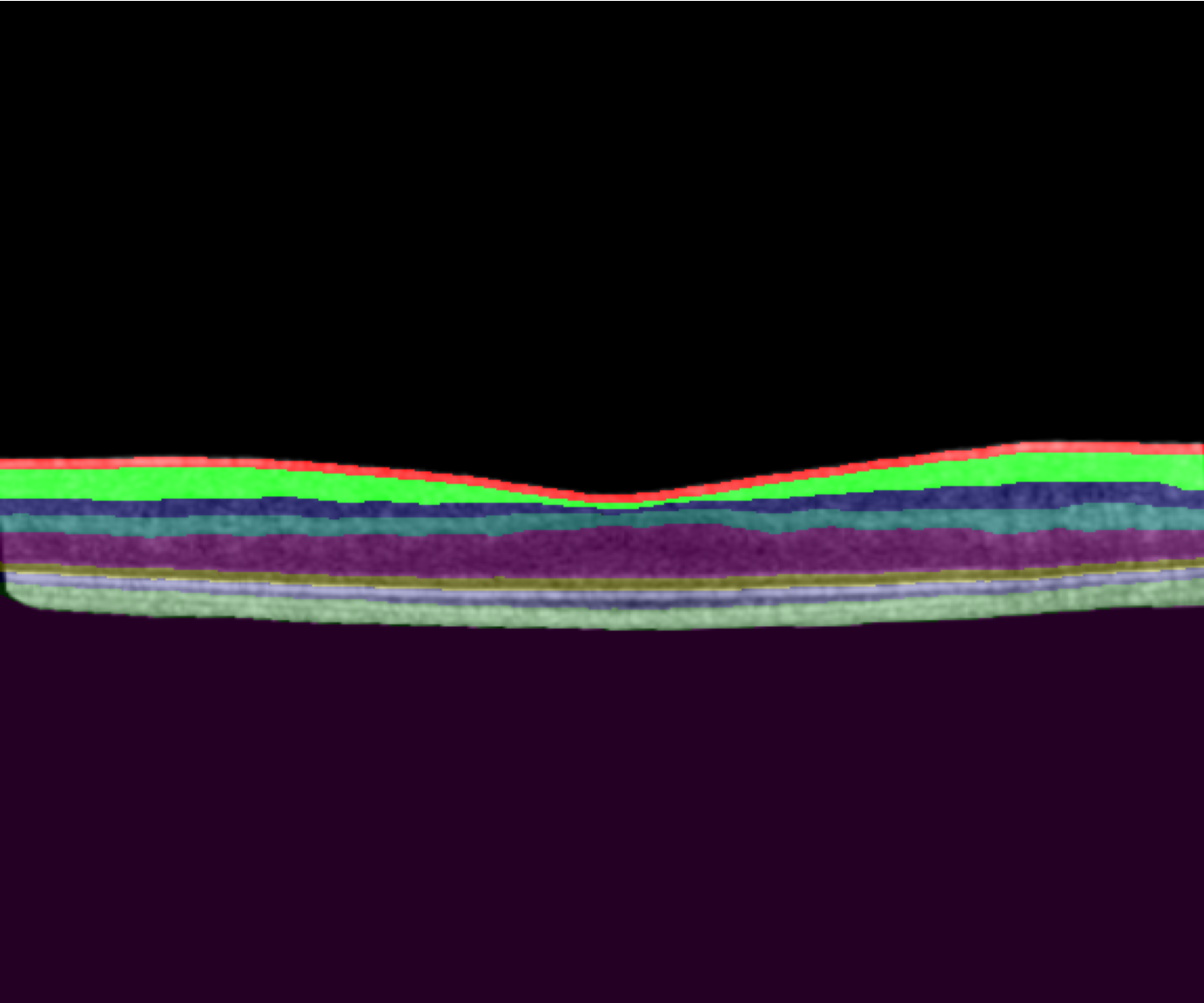}
        \caption{Foveal B-scan}
        \label{fig:seg2}
    \end{subfigure}
    \caption{Segmentation overlay on representative B-scans. The target layers for our classification task, RNFL (red) and GCL-IPL (green), are highlighted with maximum opacity, while the remaining non-target retinal layers are rendered semi-transparently to preserve anatomical context.}
    \label{fig:layers}
\end{figure}

For our classification experiments, we used the Retinal Nerve Fiber Layer (RNFL) and the Ganglion Cell Layer-Inner Plexiform Layer complex (GCL-IPL). These structures capture early neurodegenerative changes, comprising both ganglion cell axons (RNFL) and somas (GCL-IPL). Additionally, apart from evaluating each layer individually, we also concatenated their corresponding maps into multi-channel inputs to assess whether joint representation can offer complementary diagnostic information (Figure~\ref{fig:layers}).

\subsection{Models \& Architectures for Classification}

We evaluated a diverse set of machine learning and deep learning models to analyze how retinal thickness maps can be exploited by these models for ocular disease classification. Table \ref{tab:models} summarizes the different models and architectures explored in this work, ranging from tradditinal machine learning baselines to more complex deep learning architectures and hybrid approaches.

\begin{table}[t]
    \centering
    \scriptsize
    \renewcommand{\arraystretch}{1.15}
    \setlength{\tabcolsep}{4pt}
    \caption{Classification models and architectures explored in this study, organized by baseline types, autoencoder-based representations, and foundation model strategies.}
    \label{tab:models}
    \begin{tabularx}{\columnwidth}{>{\raggedright\arraybackslash}p{0.38\columnwidth} >{\raggedright\arraybackslash}X}
        \toprule
        \textbf{Model} & \textbf{Description} \\
        \midrule

        \multicolumn{2}{l}{\textit{\textbf{Tabular \& Spatial Baselines}}} \\
        \cmidrule(lr){1-2}
        XGBoost (Baseline) & Gradient Boosting Decision Trees trained on raw sector-wise thickness statistics. \\
        CNN (end-to-end) & Lightweight CNN trained end-to-end directly on 2D en-face thickness maps. \\
        \addlinespace

        \multicolumn{2}{l}{\textit{\textbf{Autoencoder-based Representations}}} \\
        \cmidrule(lr){1-2}
        VAE + MLP & Latent features extracted by a standard VAE, classified via an MLP. \\
        VAE + MLP (DINO Loss) & VAE trained using a DINO-based perceptual loss ($\mathcal{L}_{\text{DINO}}$), followed by an MLP. \\
        \addlinespace

        \multicolumn{2}{l}{\textit{\textbf{Pretrained Foundation Models (DINO Backbone)}}} \\
        \cmidrule(lr){1-2}
        DINO (Frozen) + XGBoost & Decoupled pipeline using pooled multi-layer DINO features fused with mean thickness statistics for the final XGBoost classification. \\
        DINO (Frozen) + CNN & Multi-layer 2D spatial feature maps from frozen DINO processed through a custom CNN head with a final MLP classifier. \\
        DINO (Frozen) + MLP & Fully frozen DINO backbone with pooled final-layer representations classified by a MLP head. \\
        DINO + MLP (LoRA Fine-Tuning) & End-to-end classification fine-tuning DINO's attention layers via LoRA adapters alongside the MLP head. \\
        \bottomrule
    \end{tabularx}
\end{table}

\subsubsection{Traditional Tabular and Lightweight CNN Models}

\begin{figure*}[t]
    \centering
    % Use subfigures with top alignment to keep images and captions aligned
    \begin{subfigure}[t]{0.49\linewidth}
        \centering
        \includegraphics[width=\linewidth]{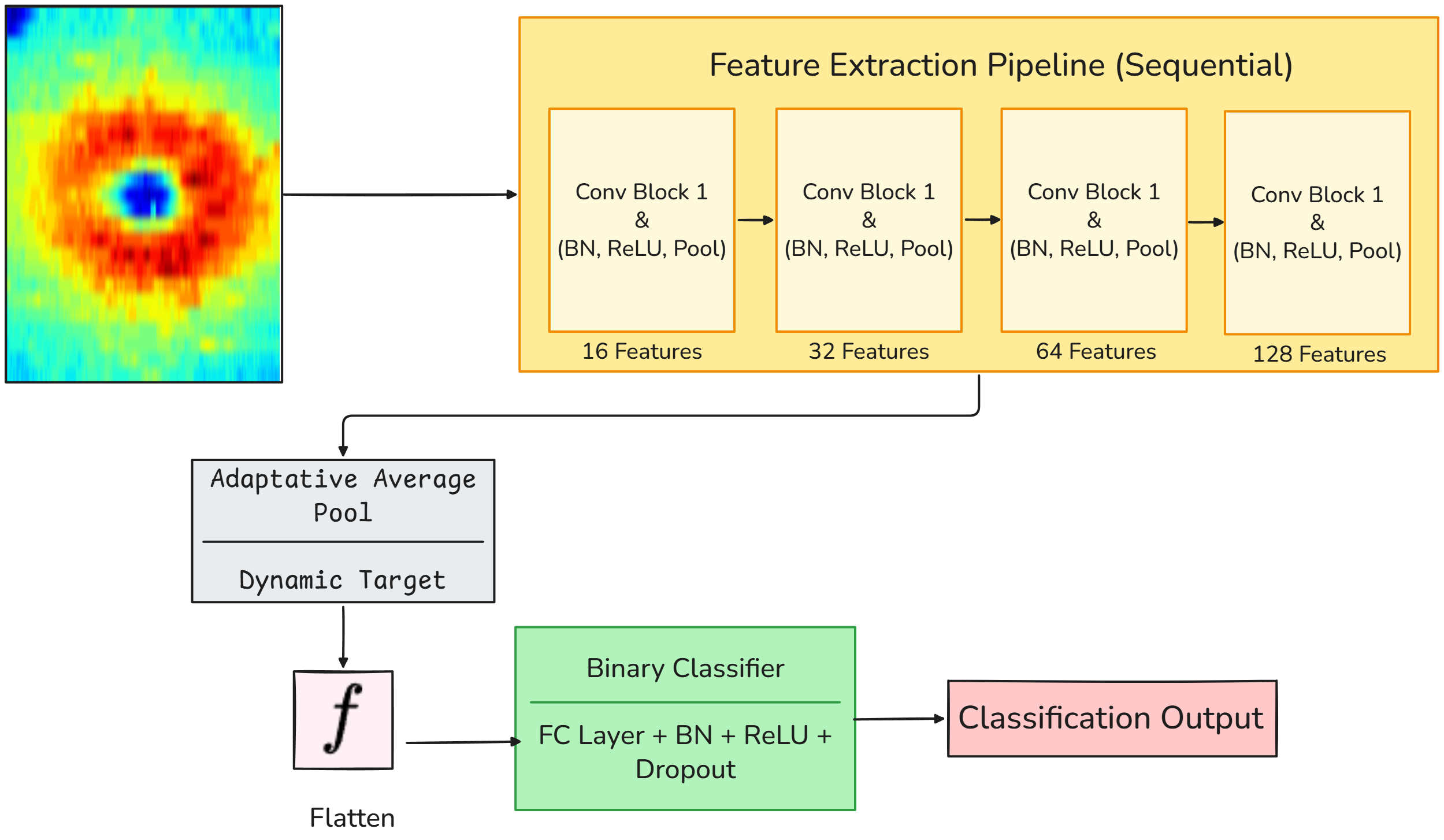}
        \caption{CNN architecture.}
        \label{fig:cnn}
    \end{subfigure}
    \hfill
    \begin{subfigure}[t]{0.49\linewidth}
        \centering
        \includegraphics[width=\linewidth]{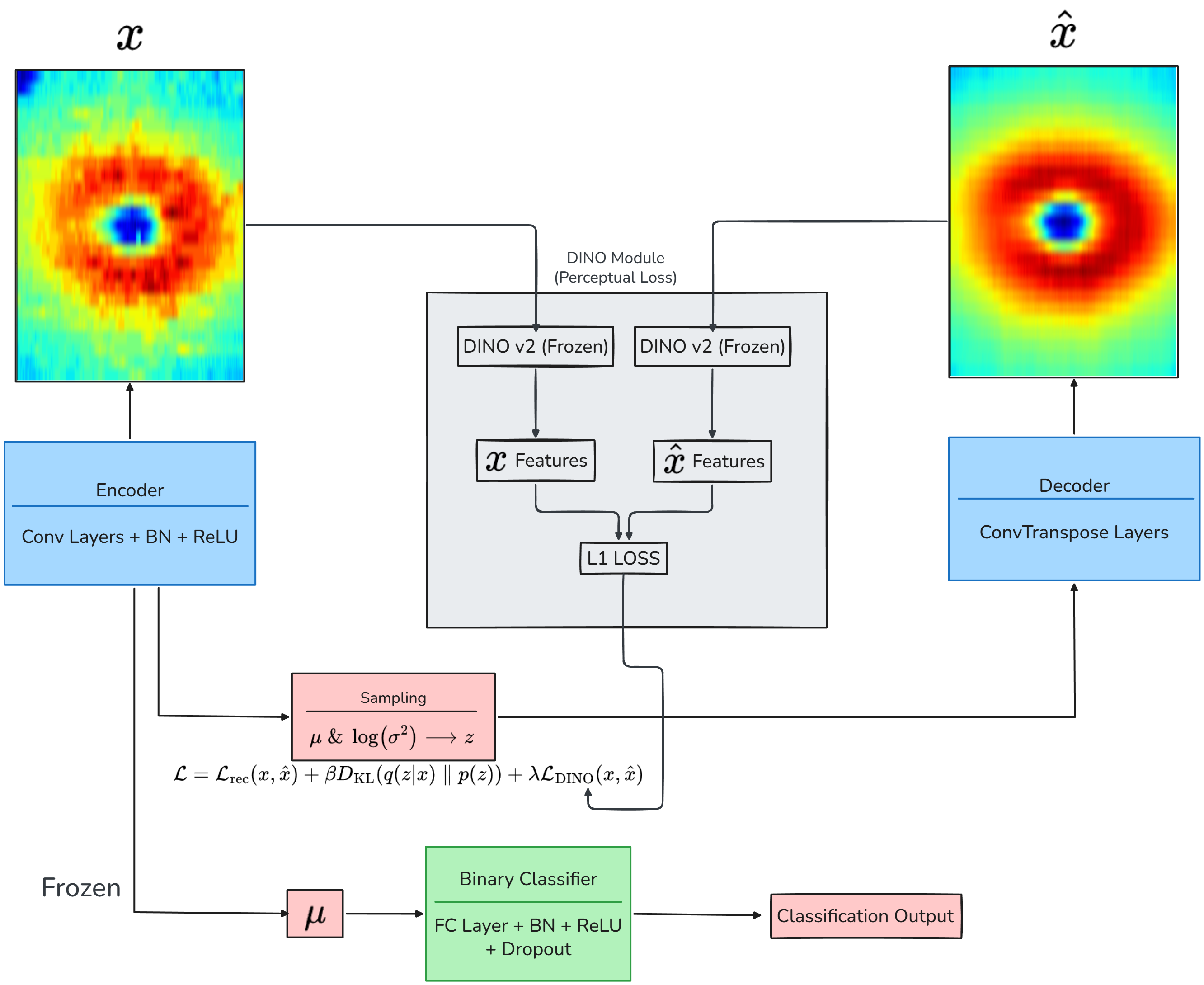}
        \caption{VAE + MLP architecture.}
        \label{fig:vae}
    \end{subfigure}
    \caption{Overview of the deep learning architectures. Left, the end-to-end CNN extracts spatial features directly from the en-face thickness maps. Right, the VAE learns a compact generative latent representation, which is subsequently processed by an MLP classifier for disease discrimination.}
    \label{fig:architectures}
\end{figure*}

We first trained an XGBoost model~\cite{chen2016xgboost} on  raw thickness values extracted from the en-face maps in tabular format. This baseline evaluates whether global, statistical summary
metrics are discriminative enough for retinal classification without spatial knowledge.

Additionally, focusing on the local spatial context, we implemented a custom lightweight Convolutional Neural Network (CNN)~\cite{o2015introduction} (Figure~\ref{fig:cnn}). The backbone consists of four sequential convolutional blocks with increasing channel capacities ($16$,$32$,$64$,$128$), each paired with Batch Normalization, ReLU activations, and max pooling for progressive spatial downsampling. Given the limited size of the available cohorts, we intentionally adopted a lightweight architecture instead of heavily parameterized networks such as ResNet or EfficientNet, reducing the risk of overfitting while preserving the ability to learn local spatial patterns. This model serves as a convolutional baseline for comparison with more complex architectures.

\subsubsection{Variational Autoencoder (VAE) for Feature Extraction}

For the unsupervised representation learning, we trained a Variational Autoencoder (VAE)~\cite{diederik2019introduction} to project retinal thickness maps into a dense N-dimensional latent space in order to capture structural patterns present in retinal thickness maps. For classification, we incorporated a Multi-Layer Perceptron (MLP)~\cite{popescu2009multilayer} head to the bottleneck
layer (Figure~\ref{fig:vae}). This architecture had two objectives: First, train the encoder-decoder structure for the unsupervised learning. Second, after the convergence, the decoder is discarded and the frozen encoder acts as a static feature extractor for the downstream MLP classifier.

The VAE is optimized via a composite loss function combining pixel-level reconstruction fidelity, latent space regularization, and an optional self-supervised perceptual constraint,

\begin{equation}
\begin{aligned}
    \mathcal{L}_{\text{total}} &= \alpha \cdot \mathcal{L}_{rec}(x, \hat{x}) \\
    & + \beta \cdot D_{\text{KL}}\left(q(z \mid x) \parallel p(z)\right) \\
    & + \gamma \cdot \mathcal{L}_{DINO}\left(x,\hat{x}\right)
\end{aligned}
\label{eq:vae_loss}
\end{equation}

where $x$ and $\hat{x}$ denote the input and the reconstructed thickness maps, $q(z|x)$ is the approximate posterior distribution of the latent variable $z$ given the input, and $p(z)$ is the prior distribution of the latent variable. The reconstruction term $\mathcal{L}_{rec}$ quantifies pixel-level fidelity using the Mean Squared Error (MSE), while the Kullback-Leibler divergence $D_{\text{KL}}$ acts as a latent regularizer forcing the approximate posterior to conform to the prior distribution. The perceptual term $\mathcal{L}_{\text{DINO}}$ measures the Mean Absolute Error (MAE) between the intermediate DINO feature representations of $x$ and $\hat{x}$. The hyperparameters $\alpha$, $\beta$, and $\gamma$ control the relative importance of each loss term during training. In the case of not using the perceptual loss, we simply set $\gamma$ to zero, which reduces the loss function to the standard VAE loss, allowing us to evaluate the impact of incorporating self-supervised feature extraction into the reconstruction loss.

\subsubsection{Pretrained Foundation Models (DINO Backbone)}

Finally, we explored the pretrained Vision Transformers (ViTs)~\cite{dosovitskiy2021an} (specifically DINOv2 and DINOv3 architectures~\cite{oquab2023dinov2,simeoni2025dinov3}) as general-purpose feature extractors. To obtain localized structural representations from retinal thickness maps (e.g., RNFL and GCL-IPL), feature extraction focuses exclusively on spatial patch tokens. Using the timm library [29], we extract representations
by enforcing an NCHW output format. This automatically discards non-spatial elements (\texttt{[CLS]} and register tokens), returning a clean spatial grid of dimensions $[B,D_{\text{embed}},H_{\text{patches}},W_{\text{patches}}]$. These spatial outputs are then routed into three different classification strategies.

\paragraph{Frozen DINO + XGBoost}

For the tree-based pipeline, feature embeddings are collected from the final $N_{\text{blocks}} = 4$ blocks of a backbone with $L$ total layers. A global average pooling is applied across the spatial dimensions of each block, giving a $D_{\text{embed}}$-dimensional vector per layer. Concatenating across these selected blocks produces a multi-scale vector of dimension $N_{\text{blocks}} \cdot D_{\text{embed}}$ per retinal structure. This multi-layer aggregation provides complementary representations extracted at different levels of the backbone, compensating for the tree model's inability to learn spatial representations via backpropagation. Finally, mean thickness metrics are appended via late fusion before passing the vector to XGBoost. Equation~\ref{eq:multilayer_features} formalizes the multi-layer feature extraction using the concatenation operator $\bigoplus$, and Equation~\ref{eq:final_features} describes the final feature vector used for classification.

\begin{equation}
\begin{aligned}
    \mathbf{f}_s = &\bigoplus_{b=L-N_{\text{blocks}}+1}^{L} \left( \frac{1}{N_{\text{patches}}} \sum_{i=1}^{N_{\text{patches}}} \mathbf{z}_{s,b}^{(i)} \right), \\
    &\quad s \in \{\text{RNFL}, \text{GCL-IPL}\}.
\end{aligned}
\label{eq:multilayer_features}
\end{equation}

\begin{equation}
\mathbf{f}_{\text{final}} = [\mathbf{f}_{\text{RNFL}} \parallel \mathbf{f}_{\text{GCL-IPL}} \parallel \mathbf{t}].
\label{eq:final_features}
\end{equation}

Here $\mathbf{z}_{s,b}^{(i)}$ represents the $D_{\text{embed}}$-dimensional feature vector of the $i$-th patch token from block $b$ for layer $s$, and $\mathbf{t}$ denotes the mean thickness metrics. The concatenation operator $\parallel$ combines the multi-layer features and thickness metrics into a single vector for XGBoost classification.

\paragraph{DINO + CNN}

\begin{figure*}[t]
    \centering
    \includegraphics[width=\linewidth]{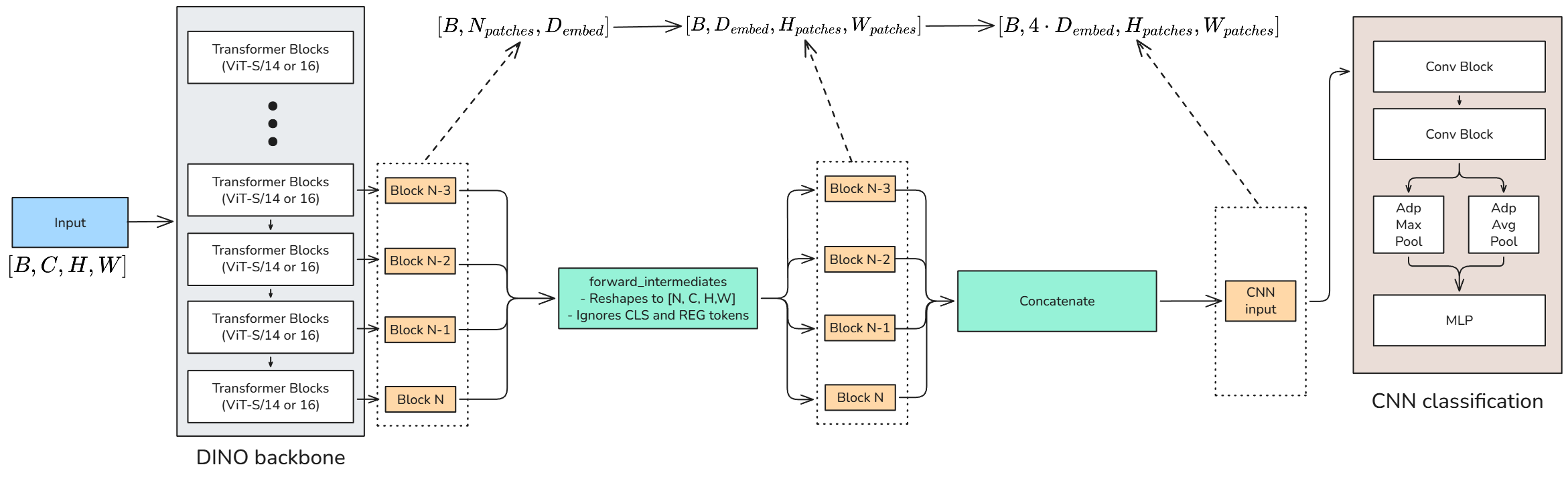}
    \caption{Architecture of the hybrid DINO + CNN classification pipeline. Intermediate patch representations are extracted from the final four transformer blocks of the DINO backbone.}
    \label{fig:dinocnn_est}
\end{figure*}

To preserve 2D spatial topographies instead of collapsing them into 1D embeddings, the hybrid DINO + CNN architecture directly processes the unpooled spatial tensors from the final $N_{\text{blocks}} = 4$ backbone blocks. Rather than depending only on the last representation layer, features from multiple blocks are concatenated along the channel dimension, allowing the network to explore multi-scale representations. This operation builds a spatial tensor of shape $[B, 4 \cdot D_{\text{embed}}, H_{\text{patches}}, W_{\text{patches}}]$. Although its spatial dimensions are smaller than those of the original input image, each spatial position contains a $4 \cdot D_{\text{embed}}$-dimensional feature vector, providing a channel-rich representation of the corresponding image region. This tensor is then processed by a convolutional head composed of two convolutional blocks each followed by Batch Normalization, ReLU activations, and max pooling for progressive spatial downsampling. The final feature map is globally pooled with an adaptive average pooling concatenated with another adaptive max pooling operation to produce a $2 \cdot D_{\text{embed}}$-dimensional vector. This vector is then forwarded to a fully connected layer for classification. Figure~\ref{fig:dinocnn_est} illustrates this hybrid architecture.

\paragraph{DINO + MLP Frozen vs. Fine-tuned with LoRA}

\begin{figure}[t]
    \centering
    \includegraphics[width=\linewidth]{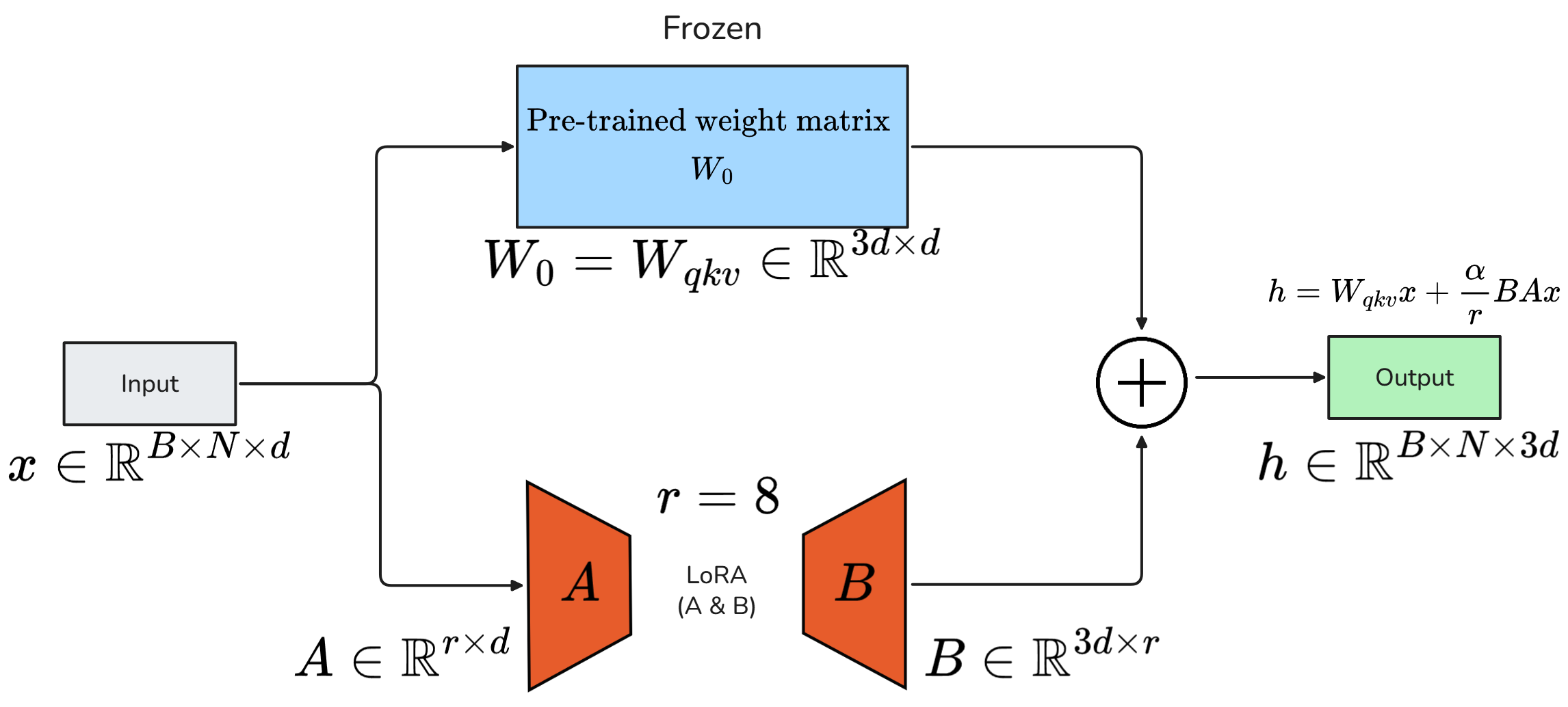}
    \caption{LoRA fine-tuning strategy for DINO-based architectures. Low-rank matrices are injected into the attention layers of the DINO backbone, allowing for efficient adaptation with minimal additional parameters. This enables the model to learn task-specific representations while retaining the generalization capabilities of the pretrained DINO features.}
    \label{fig:LoraFt}
\end{figure}

Our last method involves a fully differentiable end-to-end framework by introducing a Multi-Layer Perceptron (MLP) head. Unlike the XGBoost and CNN pipelines, this MLP receives inputs exclusively from the final transformer block ($N_{\text{blocks}} = 1$), which is globally pooled into a $D_{\text{embed}}$-dimensional vector. While the tree-based pipeline can effectively exploit concatenated representations from multiple transformer blocks ($N_{\text{blocks}} = 4$), providing a $4 \cdot D_{\text{embed}}$ input vector to a dense neural classifier would increase the number of trainable parameters in the classification head. Given the limited size of the clinical cohort, we instead keep only the final transformer block ($N_{\text{blocks}} = 1$), reducing model capacity and mitigating overfitting while preserving the most task-relevant semantic representation. This maintains a $D_{\text{embed}}$-dimensional input vector for the MLP head, ensuring a fair comparison with the following LoRA fine-tuning strategy.

With this specific setup, we evaluated two different approaches: a static, frozen backbone to establish a dimensionally aligned control baseline for raw DINO features, and implementing task-specific adaptation via Parameter-Efficient Fine-Tuning (PEFT) using Low-Rank Adaptation (LoRA)~\cite{hu2021loralowrankadaptationlarge} in order to avoid the risk of overfitting due to the data scarcity.

To adapt the pretrained representations to the downstream classification task without high computational overhead, trainable low-rank decomposition matrices ($A \in \mathbb{R}^{r \times d}, B \in \mathbb{R}^{3d \times r}$ with intrinsic rank $r = 8$) are injected directly into the query, key, and value (qkv) attention projections (Figure 9). This allows a task-specific adaptation of the attention layers while keeping the majority of the pretrained backbone parameters frozen, converting the traditional fine-tuning equation $W = W_0 + \Delta W$ into a low-rank decomposition $W = W_0 + BA$, where $W_0$ is the frozen pretrained weight matrix and $\Delta W = BA$ is the trainable low-rank update.

Together, these architectures allow us to compare consistently different strategies within the same scope, such as thickness-derived biomarkers, convolutional spatial learning, frozen self-supervised representations, and parameter-efficient adaptation of large-scale vision foundation models.

\subsection{Experimental design and training strategy}

\begin{figure}[t]
    \centering
    \includegraphics[width=\linewidth]{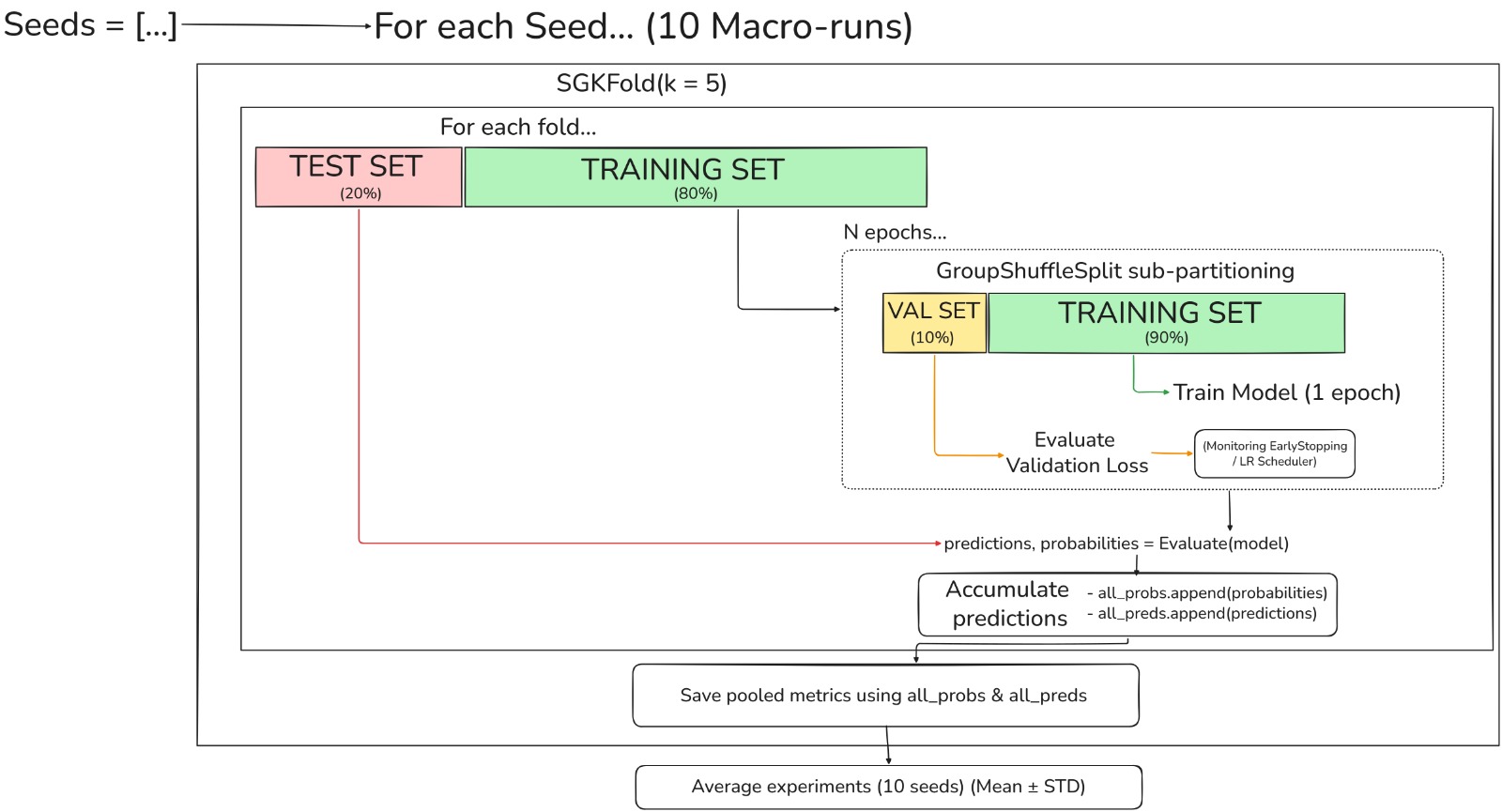}
    \caption{Stratified Group 5-fold Cross-Validation strategy across 10 independent macro-runs. The outer loop isolates an independent Test set ($20\%$), while the inner loop partitions the remaining subjects into Training ($72\%$) and Validation ($8\%$) sets, ensuring strict patient-wise grouping and class balance.}
    \label{fig:crossval}
\end{figure}

To comprehensively evaluate model performance, reproducibility, and stability, we implemented a 5-fold Stratified Group Cross-Validation strategy~\cite{prusty2022skcv} evaluated across 10 independent macro-runs with different random seeds.

To strictly prevent data leakage between splits, both the outer and inner partitions group samples by subject identifier. Because each patient in our dataset contributes images from both eyes from their most recent clinical visit, grouping by subject ensures that both eyes of any given patient are strictly assigned together to the same split (Training, Validation, or Test) and are never distributed across different sets. For each fold, a three-way partition is established as follows:

\begin{itemize}
    \item \textbf{Outer Loop (Test Set):} An outer \texttt{StratifiedGroupKFold} ($k=5$) isolates an independent Test set comprising $20\%$ of the total subject cohort.
    \item \textbf{Inner Loop (Train/Validation Sets):} The remaining $80\%$ of the data is further partitioned into Training ($72\%$ of total subjects) and Validation ($8\%$ of total subjects, corresponding to a $90\%/10\%$ relative split) using \texttt{GroupShuffleSplit} with patient-level grouping to ensure that both eyes from the same subject remain within the same subset, preventing inter-eye data leakage. The validation set is then used to monitor model convergence and guide early stopping.
\end{itemize}

To counteract class imbalance across all evaluated models, loss function class weights are dynamically computed exclusively from each fold's training distribution. Final model performance is reported using pooled metrics by accumulating predictions on independent test sets across all folds, presenting results as the mean $\pm$ standard deviation across the 10 macro-runs. Figure~\ref{fig:crossval} illustrates this cross-validation pipeline.

Additionally, all fold-dependent normalization parameters were computed exclusively from the training subset of each fold and then applied unchanged to the corresponding validation and test subsets. CNN, VAE, and DINO-CNN pipelines used channel-wise Z-score normalization based on the training-fold mean and standard deviation. In contrast, the DINO-based feature-extraction pipelines using XGBoost or an MLP applied per-sample Min-Max normalization to each en-face map, independently mapping its values to the $[0,1]$ range. These pipelines also retained raw mean-thickness statistics as additional features where applicable. The tree-based baseline used raw thickness maps without feature scaling, as tree-based models generally do not require standardized inputs. 

\subsection{Evaluation metrics}

\subsubsection{Performance Metrics}

To evaluate the performance and stability of the proposed models, we report a comprehensive set of evaluation metrics across our 5-fold cross-validation scheme over 10 independent macro-runs.

While Accuracy is reported for completeness, it may provide an overly optimistic assessment in class-imbalanced settings. Therefore, model performance is primarily evaluated using the Area Under the Receiver Operating Characteristic Curve (AUC-ROC), which measures discrimination across varying decision thresholds, and Average Precision (AP), summarizing the precision-recall curve, which is particularly informative in imbalanced medical classification tasks.

% As formalized in Equation 4, the $F_1$-score represents the harmonic mean of Precision and Recall,

% \begin{equation}
% F_1 = 2 \times \frac{\text{Precision} \times \text{Recall}}{\text{Precision} + \text{Recall}},
% \label{eq:f1_score}
% \end{equation}

% where Precision denotes the proportion of true positive predictions among all positive calls, and Recall measures the proportion of actual positive cases correctly detected. 

Furthermore, to ensure a balanced evaluation of class-specific behavior without bias toward the majority class, we report the class-wise $F_1$-scores as the mean $\pm$ standard deviation across all evaluations. 

\subsubsection{Statistical Significance Analysis}

To determine which performances are statistically significant, we implemented a non-parametric statistical framework across the 10 independent macro-runs.

Specifically, we perform pairwise comparisons using the non-parametric Wilcoxon signed-rank test [32] applied to the performance scores obtained from the 10 independent macro-runs. To evaluate statistical dominance, a many-to-one testing scheme is applied, comparing the best-performing model according to the target metric against each remaining architecture individually. Additionally, to control the False Discovery Rate (FDR) arising from multiple testing, raw $p$-values are adjusted using the Benjamini-Hochberg procedure~\cite{benjamini1995controlling} at a significance threshold of $q = 0.05$.

In our reporting, the top-performing model along with any models demonstrating statistical equivalence ($p_{\text{adj}} \ge 0.05$) to it are highlighted in \textbf{bold}. Furthermore, the top model is marked with a dagger ($\dagger$) if and only if it achieves strict statistical superiority ($p_{\text{adj}} < 0.05$) over all other evaluated architectures.

\subsection{Explainability and Interpretability}

Beyond quantitative performance evaluation, qualitative assessment is also important in medical imaging applications, as it helps to explore whether model decisions are associated with anatomically meaningful regions. In this context, the high-resolution en-face thickness maps derived from retinal layer segmentations offer a spatially detailed representation of retinal morphology, unlike conventional regional thickness measurements that average information over predefined sectors. This spatial detail enables a more localized analysis of the regions contributing to model predictions, potentially revealing patterns that would otherwise be obscured by regional averaging. To investigate these spatial contributions, we employed two complementary interpretability methods, SHAP~\cite{shap} and Grad-CAM++~\cite{gradcam}.

SHAP is a unified explainability framework based on cooperative game theory that attributes feature importance according to their contribution to the model prediction. In our study, SHAP was used to estimate the contribution of individual pixels to the final classification decision, providing fine-grained pixel-level explanations. For tree-based models, explanations were computed using TreeExplainer, whereas GradientExplainer was used for differentiable neural architectures. The main drawback of SHAP is its computational cost, particularly when estimating attribution maps for high-dimensional image inputs. Furthermore, SHAP inherently relies on a reference dataset (background set) that acts as a neutral state to contrast prediction deviations. In our work, in order to optimize computation time without compromising the stability of the explanations, we established a background set of 50 images as a neutral reference point. These reference images were randomly sampled exclusively from the training dataset to prevent data leakage and ensure that the baseline accurately represents the distribution learned by the model.

On the other hand, we have used Grad-CAM++~\cite{chattopadhay2018grad}, an extension of the original Grad-CAM method, which provides visual explanations for differentiable vision models by highlighting the regions in the input image that are most influential for a given prediction. It improves the original method by considering higher-order derivatives, providing more localized attribution maps than the original Grad-CAM formulation. In our case, Grad-CAM++ was used to generate heatmaps highlighting the regions of the en-face maps that were most influential for the classification decisions of the evaluated end-to-end neural architectures. SHAP and Grad-CAM++ provide complementary perspectives on model interpretability. SHAP generates fine-grained attribution maps that can indicate both positive and negative contributions to a prediction. In contrast, Grad-CAM++ highlights regions that are influential for a given prediction but does not directly encode the sign of the contribution.

\subsection{Hyperparameter Optimization}

\begin{table}[t]
\centering
\footnotesize
\caption{Main hyperparameter configurations used in the experiments.}
\label{tab:hyperparameters}
\begin{tabular}{@{} l p{0.62\columnwidth} @{}}
\toprule
\textbf{Model} & \textbf{Main hyperparameters} \\ 
\midrule
XGBoost (Baseline) & $\text{Depth} = 3$, $\text{learning rate} = 0.005$ \\
\addlinespace
CNN (end-to-end) & Conv channels: $(16, 32, 64, 128)$, $\text{dropout} = 0.3$ \\
\addlinespace
VAE + MLP & $\text{Latent dim} = 128$, $\alpha = 5$, $\beta = 1$, $\gamma = 0.5$ (DINO variant) \\
\addlinespace
DINO (Frozen) + XGBoost & Final $4$ blocks, $\text{depth} = 3$, $\text{learning rate} = 0.01$ \\
\addlinespace
DINO (Frozen) + CNN & Final $4$ blocks, two-block CNN head \\
\addlinespace
DINO (Frozen) + MLP & Final layer, $\text{hidden dim} = 128$, $\text{dropout} = 0.5$ \\
\addlinespace
DINO + MLP (LoRA) & $qkv$ adaptation, $\text{rank } r = 8$, $\text{scaling } \alpha = 16$, $\text{dropout} = 0.05$ \\ 
\bottomrule
\end{tabular}
\end{table}

Hyperparameters were optimized independently within each training fold. Given the computational cost of some architectures, a limited set of candidate configurations was evaluated for each model based on preliminary experiments and commonly used settings from the literature. The final configuration was selected according to validation performance. Table~\ref{tab:hyperparameters} summarizes the main values used in the reported experiments. All models used early stopping, learning-rate scheduling, and weight decay regularization. The maximum training duration was adapted to each architecture depending on the complexity of the model and the risk of overfitting.

\section{Results}
\label{sec:results}

This section presents the results of the experiments conducted to evaluate the performance of different machine learning methods on the FastMacula dataset for the classification of healthy controls (HC) against different pathological conditions, including multiple sclerosis (MS), Parkinson's disease (PD), and Alzheimer's disease (AD).
\begin{table*}[t]
\centering
\caption{Comprehensive performance comparison across different retinal layer configurations (RNFL, GCL-IPL, and RNFL + GCL-IPL) for the HC vs. MS, PD, and AD classification tasks. Values are reported as mean $\pm$ standard deviation ($\sigma, N = 10$). Bold values indicate models within the top-performing statistical tier according to a post-hoc Wilcoxon signed-rank test ($p_{\text{adj}} \geq 0.05$).}
\label{tab:comprehensive_all_datasets}
\resizebox{\textwidth}{!}{%
\begin{tabular}{l ccccc c ccccc c ccccc}
\toprule
& \multicolumn{5}{c}{\textbf{RNFL}} & & \multicolumn{5}{c}{\textbf{GCL-IPL}} & & \multicolumn{5}{c}{\textbf{RNFL + GCL-IPL}} \\
\cmidrule(lr){2-6} \cmidrule(lr){8-12} \cmidrule(lr){14-18}
Model & AUC & AP & Acc. & $F_1$-HC & $F_1$-Dataset & & AUC & AP & Acc. & $F_1$-HC & $F_1$-Dataset & & AUC & AP & Acc. & $F_1$-HC & $F_1$-Dataset \\
\midrule
\multicolumn{18}{l}{\textit{\textbf{Dataset: MS (Multiple Sclerosis)}}} \\
\midrule
CNN (end-to-end) & \textbf{0.81 $\pm$ 0.01} & \textbf{0.92 $\pm$ 0.00} & \textbf{0.75 $\pm$ 0.01} & \textbf{0.60 $\pm$ 0.03} & \textbf{0.82 $\pm$ 0.01} & & \textbf{0.81 $\pm$ 0.01} & 0.92 $\pm$ 0.01 & \textbf{0.74 $\pm$ 0.02} & \textbf{0.60 $\pm$ 0.03} & \textbf{0.80 $\pm$ 0.02} & & \textbf{0.82 $\pm$ 0.01} & \textbf{0.93 $\pm$ 0.01} & \textbf{0.74 $\pm$ 0.02} & \textbf{0.61 $\pm$ 0.02} & \textbf{0.80 $\pm$ 0.02} \\
VAE + MLP & \textbf{0.81 $\pm$ 0.01} & \textbf{0.92 $\pm$ 0.01} & 0.73 $\pm$ 0.01 & \textbf{0.60 $\pm$ 0.02} & 0.80 $\pm$ 0.01 & & \textbf{0.82 $\pm$ 0.01} & \textbf{0.93 $\pm$ 0.00} & \textbf{0.74 $\pm$ 0.02} & \textbf{0.62 $\pm$ 0.02} & \textbf{0.80 $\pm$ 0.01} & & \textbf{0.82 $\pm$ 0.01} & \textbf{0.93 $\pm$ 0.00} & \textbf{0.73 $\pm$ 0.01} & \textbf{0.60 $\pm$ 0.03} & 0.79 $\pm$ 0.01 \\
VAE + MLP (w\ DINO v2) & \textbf{0.81 $\pm$ 0.01} & \textbf{0.92 $\pm$ 0.00} & 0.72 $\pm$ 0.01 & \textbf{0.60 $\pm$ 0.01} & 0.79 $\pm$ 0.01 & & \textbf{0.82 $\pm$ 0.02} & \textbf{0.93 $\pm$ 0.01} & \textbf{0.73 $\pm$ 0.01} & \textbf{0.62 $\pm$ 0.02} & \textbf{0.80 $\pm$ 0.01} & & \textbf{0.82 $\pm$ 0.01} & \textbf{0.93 $\pm$ 0.00} & \textbf{0.73 $\pm$ 0.01} & \textbf{0.61 $\pm$ 0.02} & 0.79 $\pm$ 0.01 \\
VAE + MLP (w\ DINO v3) & \textbf{0.81 $\pm$ 0.01} & \textbf{0.92 $\pm$ 0.00} & 0.72 $\pm$ 0.01 & \textbf{0.60 $\pm$ 0.01} & 0.79 $\pm$ 0.01 & & \textbf{0.81 $\pm$ 0.01} & \textbf{0.93 $\pm$ 0.00} & \textbf{0.73 $\pm$ 0.01} & \textbf{0.62 $\pm$ 0.01} & \textbf{0.79 $\pm$ 0.01} & & \textbf{0.82 $\pm$ 0.01} & \textbf{0.93 $\pm$ 0.00} & \textbf{0.72 $\pm$ 0.02} & \textbf{0.60 $\pm$ 0.02} & 0.79 $\pm$ 0.01 \\
DINO v2 + XGB & 0.78 $\pm$ 0.01 & 0.90 $\pm$ 0.01 & \textbf{0.74 $\pm$ 0.01} & 0.58 $\pm$ 0.02 & \textbf{0.81 $\pm$ 0.01} & & 0.77 $\pm$ 0.01 & 0.90 $\pm$ 0.01 & 0.70 $\pm$ 0.01 & 0.55 $\pm$ 0.02 & 0.78 $\pm$ 0.01 & & 0.80 $\pm$ 0.01 & 0.91 $\pm$ 0.01 & \textbf{0.73 $\pm$ 0.01} & 0.57 $\pm$ 0.01 & \textbf{0.80 $\pm$ 0.01} \\
DINO v3 + XGB & 0.78 $\pm$ 0.01 & 0.90 $\pm$ 0.01 & 0.72 $\pm$ 0.01 & 0.56 $\pm$ 0.01 & 0.80 $\pm$ 0.01 & & 0.77 $\pm$ 0.02 & 0.90 $\pm$ 0.01 & 0.71 $\pm$ 0.01 & 0.56 $\pm$ 0.02 & \textbf{0.78 $\pm$ 0.01} & & 0.79 $\pm$ 0.01 & 0.91 $\pm$ 0.01 & \textbf{0.73 $\pm$ 0.01} & 0.58 $\pm$ 0.01 & \textbf{0.80 $\pm$ 0.01} \\
DINO v2 + CNN & 0.72 $\pm$ 0.03 & 0.87 $\pm$ 0.02 & 0.70 $\pm$ 0.03 & 0.47 $\pm$ 0.05 & 0.79 $\pm$ 0.04 & & 0.75 $\pm$ 0.02 & 0.89 $\pm$ 0.01 & 0.71 $\pm$ 0.02 & 0.48 $\pm$ 0.05 & \textbf{0.79 $\pm$ 0.02} & & 0.75 $\pm$ 0.02 & 0.89 $\pm$ 0.01 & \textbf{0.71 $\pm$ 0.01} & 0.48 $\pm$ 0.06 & 0.79 $\pm$ 0.01 \\
DINO v3 + CNN & 0.72 $\pm$ 0.01 & 0.87 $\pm$ 0.01 & 0.70 $\pm$ 0.02 & 0.45 $\pm$ 0.04 & 0.79 $\pm$ 0.02 & & 0.78 $\pm$ 0.02 & 0.91 $\pm$ 0.01 & \textbf{0.72 $\pm$ 0.02} & 0.52 $\pm$ 0.05 & \textbf{0.81 $\pm$ 0.02} & & 0.75 $\pm$ 0.02 & 0.89 $\pm$ 0.02 & 0.70 $\pm$ 0.01 & 0.48 $\pm$ 0.06 & 0.79 $\pm$ 0.02 \\
DINO v2 + MLP \textbar{} FT & 0.75 $\pm$ 0.01 & 0.88 $\pm$ 0.01 & 0.70 $\pm$ 0.01 & 0.53 $\pm$ 0.03 & 0.78 $\pm$ 0.01 & & 0.74 $\pm$ 0.01 & 0.89 $\pm$ 0.01 & 0.69 $\pm$ 0.02 & 0.51 $\pm$ 0.02 & 0.78 $\pm$ 0.02 & & 0.76 $\pm$ 0.01 & 0.90 $\pm$ 0.01 & \textbf{0.71 $\pm$ 0.02} & 0.52 $\pm$ 0.03 & 0.79 $\pm$ 0.02 \\
DINO v3 + MLP \textbar{} FT & 0.77 $\pm$ 0.01 & 0.90 $\pm$ 0.01 & 0.72 $\pm$ 0.01 & 0.52 $\pm$ 0.01 & 0.80 $\pm$ 0.01 & & 0.77 $\pm$ 0.02 & 0.91 $\pm$ 0.00 & 0.72 $\pm$ 0.02 & 0.54 $\pm$ 0.03 & \textbf{0.80 $\pm$ 0.02} & & 0.79 $\pm$ 0.01 & 0.91 $\pm$ 0.01 & \textbf{0.74 $\pm$ 0.01} & 0.56 $\pm$ 0.03 & \textbf{0.81 $\pm$ 0.01} \\
DINO v2 + MLP \textbar{} no FT & 0.76 $\pm$ 0.01 & 0.89 $\pm$ 0.01 & 0.71 $\pm$ 0.02 & 0.55 $\pm$ 0.03 & 0.79 $\pm$ 0.01 & & 0.74 $\pm$ 0.01 & 0.89 $\pm$ 0.01 & 0.68 $\pm$ 0.02 & 0.50 $\pm$ 0.03 & 0.77 $\pm$ 0.02 & & 0.75 $\pm$ 0.01 & 0.90 $\pm$ 0.01 & 0.69 $\pm$ 0.02 & 0.50 $\pm$ 0.02 & 0.78 $\pm$ 0.02 \\
DINO v3 + MLP \textbar{} no FT & 0.77 $\pm$ 0.01 & 0.90 $\pm$ 0.01 & 0.71 $\pm$ 0.01 & 0.53 $\pm$ 0.02 & 0.79 $\pm$ 0.01 & & 0.78 $\pm$ 0.02 & 0.91 $\pm$ 0.01 & 0.72 $\pm$ 0.02 & 0.53 $\pm$ 0.03 & \textbf{0.80 $\pm$ 0.02} & & 0.79 $\pm$ 0.01 & 0.91 $\pm$ 0.01 & \textbf{0.74 $\pm$ 0.01} & 0.55 $\pm$ 0.03 & \textbf{0.81 $\pm$ 0.01} \\

\midrule
\multicolumn{18}{l}{\textit{\textbf{Dataset: PD (Parkinson's Disease)}}} \\
\midrule

CNN (end-to-end) & 0.60 $\pm$ 0.02 & 0.68 $\pm$ 0.02 & \textbf{0.57 $\pm$ 0.02} & 0.46 $\pm$ 0.07 & \textbf{0.64 $\pm$ 0.05} & & \textbf{0.73 $\pm$ 0.02} & \textbf{0.78 $\pm$ 0.01} & \textbf{0.68 $\pm$ 0.02} & \textbf{0.60 $\pm$ 0.05} & \textbf{0.73 $\pm$ 0.03} & & 0.71 $\pm$ 0.03 & 0.74 $\pm$ 0.03 & 0.66 $\pm$ 0.03 & 0.59 $\pm$ 0.04 & \textbf{0.70 $\pm$ 0.04} \\
VAE + MLP & 0.58 $\pm$ 0.03 & 0.66 $\pm$ 0.02 & 0.55 $\pm$ 0.02 & 0.51 $\pm$ 0.03 & 0.58 $\pm$ 0.03 & & \textbf{0.73 $\pm$ 0.02} & \textbf{0.76 $\pm$ 0.02} & \textbf{0.67 $\pm$ 0.02} & \textbf{0.62 $\pm$ 0.02} & \textbf{0.70 $\pm$ 0.02} & & 0.70 $\pm$ 0.02 & 0.73 $\pm$ 0.02 & \textbf{0.66 $\pm$ 0.01} & \textbf{0.62 $\pm$ 0.02} & \textbf{0.69 $\pm$ 0.02} \\
VAE + MLP (w\ DINO v2) & 0.58 $\pm$ 0.02 & 0.67 $\pm$ 0.02 & 0.56 $\pm$ 0.02 & 0.52 $\pm$ 0.04 & 0.59 $\pm$ 0.02 & & \textbf{0.73 $\pm$ 0.02} & \textbf{0.76 $\pm$ 0.02} & \textbf{0.66 $\pm$ 0.02} & \textbf{0.62 $\pm$ 0.02} & 0.69 $\pm$ 0.02 & & 0.70 $\pm$ 0.03 & 0.73 $\pm$ 0.03 & 0.65 $\pm$ 0.02 & 0.60 $\pm$ 0.03 & \textbf{0.69 $\pm$ 0.02} \\
VAE + MLP (w\ DINO v3) & 0.59 $\pm$ 0.02 & 0.67 $\pm$ 0.02 & 0.55 $\pm$ 0.02 & 0.51 $\pm$ 0.04 & 0.59 $\pm$ 0.02 & & \textbf{0.73 $\pm$ 0.02} & \textbf{0.77 $\pm$ 0.02} & \textbf{0.67 $\pm$ 0.03} & \textbf{0.62 $\pm$ 0.04} & \textbf{0.71 $\pm$ 0.03} & & 0.69 $\pm$ 0.02 & 0.72 $\pm$ 0.03 & 0.65 $\pm$ 0.02 & 0.60 $\pm$ 0.02 & \textbf{0.69 $\pm$ 0.02} \\
DINO v2 + XGB & \textbf{0.61 $\pm$ 0.03} & 0.67 $\pm$ 0.03 & \textbf{0.58 $\pm$ 0.02} & \textbf{0.54 $\pm$ 0.03} & 0.62 $\pm$ 0.02 & & \textbf{0.72 $\pm$ 0.03} & \textbf{0.76 $\pm$ 0.03} & \textbf{0.68 $\pm$ 0.03} & \textbf{0.65 $\pm$ 0.02} & \textbf{0.71 $\pm$ 0.04} & & 0.72 $\pm$ 0.03 & 0.75 $\pm$ 0.03 & \textbf{0.68 $\pm$ 0.02} & \textbf{0.65 $\pm$ 0.02} & \textbf{0.71 $\pm$ 0.02} \\
DINO v3 + XGB & \textbf{0.62 $\pm$ 0.03} & 0.67 $\pm$ 0.03 & \textbf{0.60 $\pm$ 0.03} & \textbf{0.55 $\pm$ 0.03} & \textbf{0.63 $\pm$ 0.04} & & \textbf{0.70 $\pm$ 0.03} & 0.72 $\pm$ 0.03 & \textbf{0.67 $\pm$ 0.02} & \textbf{0.63 $\pm$ 0.02} & \textbf{0.70 $\pm$ 0.02} & & 0.69 $\pm$ 0.03 & 0.73 $\pm$ 0.03 & \textbf{0.66 $\pm$ 0.03} & 0.62 $\pm$ 0.03 & \textbf{0.69 $\pm$ 0.04} \\
DINO v2 + CNN & \textbf{0.62 $\pm$ 0.03} & 0.67 $\pm$ 0.04 & \textbf{0.59 $\pm$ 0.02} & 0.51 $\pm$ 0.05 & \textbf{0.64 $\pm$ 0.05} & & \textbf{0.73 $\pm$ 0.03} & 0.77 $\pm$ 0.03 & \textbf{0.67 $\pm$ 0.03} & \textbf{0.60 $\pm$ 0.08} & \textbf{0.71 $\pm$ 0.03} & & 0.73 $\pm$ 0.02 & \textbf{0.78 $\pm$ 0.03} & \textbf{0.68 $\pm$ 0.03} & \textbf{0.61 $\pm$ 0.05} & \textbf{0.72 $\pm$ 0.04} \\
DINO v3 + CNN & \textbf{0.61 $\pm$ 0.03} & 0.67 $\pm$ 0.03 & \textbf{0.59 $\pm$ 0.03} & 0.51 $\pm$ 0.06 & \textbf{0.64 $\pm$ 0.04} & & \textbf{0.73 $\pm$ 0.03} & \textbf{0.76 $\pm$ 0.03} & \textbf{0.68 $\pm$ 0.03} & \textbf{0.63 $\pm$ 0.05} & \textbf{0.72 $\pm$ 0.04} & & \textbf{0.75 $\pm$ 0.01}$^\dagger$ & \textbf{0.79 $\pm$ 0.01} & \textbf{0.69 $\pm$ 0.03} & \textbf{0.64 $\pm$ 0.04} & \textbf{0.72 $\pm$ 0.05} \\
DINO v2 + MLP \textbar{} FT & \textbf{0.62 $\pm$ 0.03} & 0.67 $\pm$ 0.02 & \textbf{0.60 $\pm$ 0.03} & \textbf{0.53 $\pm$ 0.05} & \textbf{0.65 $\pm$ 0.03} & & \textbf{0.70 $\pm$ 0.03} & \textbf{0.75 $\pm$ 0.02} & \textbf{0.67 $\pm$ 0.03} & \textbf{0.63 $\pm$ 0.05} & \textbf{0.70 $\pm$ 0.02} & & 0.70 $\pm$ 0.02 & 0.74 $\pm$ 0.02 & \textbf{0.67 $\pm$ 0.02} & \textbf{0.62 $\pm$ 0.03} & \textbf{0.70 $\pm$ 0.02} \\
DINO v3 + MLP \textbar{} FT & \textbf{0.63 $\pm$ 0.03} & \textbf{0.70 $\pm$ 0.02} & \textbf{0.61 $\pm$ 0.03} & \textbf{0.54 $\pm$ 0.04} & \textbf{0.65 $\pm$ 0.03} & & \textbf{0.70 $\pm$ 0.03} & \textbf{0.75 $\pm$ 0.02} & \textbf{0.65 $\pm$ 0.02} & 0.58 $\pm$ 0.05 & 0.70 $\pm$ 0.02 & & 0.71 $\pm$ 0.03 & 0.75 $\pm$ 0.03 & 0.65 $\pm$ 0.02 & 0.57 $\pm$ 0.04 & 0.70 $\pm$ 0.02 \\
DINO v2 + MLP \textbar{} no FT & 0.60 $\pm$ 0.02 & 0.66 $\pm$ 0.02 & \textbf{0.58 $\pm$ 0.02} & 0.51 $\pm$ 0.04 & 0.64 $\pm$ 0.02 & & \textbf{0.71 $\pm$ 0.02} & 0.75 $\pm$ 0.02 & \textbf{0.67 $\pm$ 0.02} & \textbf{0.63 $\pm$ 0.03} & 0.69 $\pm$ 0.02 & & 0.71 $\pm$ 0.03 & 0.75 $\pm$ 0.02 & \textbf{0.66 $\pm$ 0.02} & \textbf{0.62 $\pm$ 0.02} & 0.70 $\pm$ 0.02 \\
DINO v3 + MLP \textbar{} no FT & \textbf{0.63 $\pm$ 0.03} & \textbf{0.70 $\pm$ 0.02} & \textbf{0.61 $\pm$ 0.01} & \textbf{0.53 $\pm$ 0.04} & \textbf{0.66 $\pm$ 0.02} & & 0.68 $\pm$ 0.02 & 0.75 $\pm$ 0.01 & 0.64 $\pm$ 0.02 & \textbf{0.59 $\pm$ 0.04} & 0.67 $\pm$ 0.03 & & 0.70 $\pm$ 0.02 & 0.75 $\pm$ 0.02 & \textbf{0.65 $\pm$ 0.02} & \textbf{0.59 $\pm$ 0.03} & 0.69 $\pm$ 0.02 \\

\midrule
\multicolumn{18}{l}{\textit{\textbf{Dataset: AD (Alzheimer's Disease)}}} \\
\midrule
CNN (end-to-end) & 0.63 $\pm$ 0.04 & 0.44 $\pm$ 0.05 & 0.67 $\pm$ 0.04 & 0.77 $\pm$ 0.04 & 0.39 $\pm$ 0.08 & & \textbf{0.75 $\pm$ 0.04} & \textbf{0.63 $\pm$ 0.04} & 0.75 $\pm$ 0.03 & 0.82 $\pm$ 0.03 & \textbf{0.57 $\pm$ 0.04} & & \textbf{0.74 $\pm$ 0.04} & 0.54 $\pm$ 0.05 & \textbf{0.75 $\pm$ 0.05} & 0.83 $\pm$ 0.04 & \textbf{0.58 $\pm$ 0.06} \\
VAE + MLP & 0.60 $\pm$ 0.04 & 0.39 $\pm$ 0.04 & 0.62 $\pm$ 0.04 & 0.72 $\pm$ 0.03 & 0.41 $\pm$ 0.06 & & 0.75 $\pm$ 0.02 & \textbf{0.59 $\pm$ 0.04} & 0.76 $\pm$ 0.01 & 0.83 $\pm$ 0.01 & \textbf{0.60 $\pm$ 0.02} & & 0.73 $\pm$ 0.03 & 0.50 $\pm$ 0.03 & \textbf{0.76 $\pm$ 0.02} & \textbf{0.83 $\pm$ 0.02} & \textbf{0.60 $\pm$ 0.03} \\
VAE + MLP (w\ DINO v2) & 0.59 $\pm$ 0.04 & 0.38 $\pm$ 0.03 & 0.63 $\pm$ 0.02 & 0.73 $\pm$ 0.02 & 0.40 $\pm$ 0.05 & & 0.75 $\pm$ 0.02 & 0.58 $\pm$ 0.04 & 0.77 $\pm$ 0.01 & 0.83 $\pm$ 0.01 & \textbf{0.60 $\pm$ 0.03} & & 0.73 $\pm$ 0.03 & 0.49 $\pm$ 0.05 & 0.75 $\pm$ 0.02 & 0.82 $\pm$ 0.02 & \textbf{0.59 $\pm$ 0.04} \\
VAE + MLP (w\ DINO v3) & 0.59 $\pm$ 0.04 & 0.37 $\pm$ 0.05 & 0.61 $\pm$ 0.04 & 0.71 $\pm$ 0.04 & 0.40 $\pm$ 0.05 & & \textbf{0.76 $\pm$ 0.02} & \textbf{0.60 $\pm$ 0.05} & 0.76 $\pm$ 0.02 & 0.83 $\pm$ 0.01 & \textbf{0.60 $\pm$ 0.02} & & 0.73 $\pm$ 0.01 & 0.50 $\pm$ 0.03 & \textbf{0.76 $\pm$ 0.03} & 0.83 $\pm$ 0.02 & \textbf{0.60 $\pm$ 0.03} \\
DINO v2 + XGB & \textbf{0.69 $\pm$ 0.03} & 0.45 $\pm$ 0.03 & \textbf{0.72 $\pm$ 0.03} & \textbf{0.80 $\pm$ 0.02} & \textbf{0.50 $\pm$ 0.05} & & \textbf{0.77 $\pm$ 0.02} & \textbf{0.59 $\pm$ 0.03} & \textbf{0.78 $\pm$ 0.01} & \textbf{0.85 $\pm$ 0.01} & \textbf{0.58 $\pm$ 0.03} & & \textbf{0.77 $\pm$ 0.02} & \textbf{0.62 $\pm$ 0.03} & \textbf{0.78 $\pm$ 0.02} & \textbf{0.85 $\pm$ 0.02} & \textbf{0.58 $\pm$ 0.04} \\
DINO v3 + XGB & 0.65 $\pm$ 0.03 & 0.42 $\pm$ 0.05 & 0.68 $\pm$ 0.03 & \textbf{0.78 $\pm$ 0.02} & 0.43 $\pm$ 0.04 & & 0.74 $\pm$ 0.01 & \textbf{0.58 $\pm$ 0.04} & 0.76 $\pm$ 0.02 & 0.83 $\pm$ 0.02 & 0.55 $\pm$ 0.03 & & 0.74 $\pm$ 0.01 & 0.59 $\pm$ 0.03 & \textbf{0.76 $\pm$ 0.02} & \textbf{0.84 $\pm$ 0.01} & 0.54 $\pm$ 0.04 \\
DINO v2 + CNN & 0.62 $\pm$ 0.03 & 0.41 $\pm$ 0.03 & \textbf{0.70 $\pm$ 0.02} & \textbf{0.80 $\pm$ 0.02} & 0.37 $\pm$ 0.04 & & 0.74 $\pm$ 0.03 & 0.57 $\pm$ 0.04 & 0.76 $\pm$ 0.02 & 0.84 $\pm$ 0.01 & 0.49 $\pm$ 0.03 & & \textbf{0.75 $\pm$ 0.02} & \textbf{0.63 $\pm$ 0.03} & \textbf{0.78 $\pm$ 0.02} & \textbf{0.85 $\pm$ 0.02} & 0.55 $\pm$ 0.04 \\
DINO v3 + CNN & \textbf{0.67 $\pm$ 0.02} & \textbf{0.51 $\pm$ 0.02}$^\dagger$ & \textbf{0.71 $\pm$ 0.03} & \textbf{0.81 $\pm$ 0.03} & 0.44 $\pm$ 0.03 & & 0.67 $\pm$ 0.04 & 0.51 $\pm$ 0.06 & 0.73 $\pm$ 0.03 & 0.82 $\pm$ 0.03 & 0.46 $\pm$ 0.04 & & \textbf{0.76 $\pm$ 0.03} & \textbf{0.63 $\pm$ 0.04} & \textbf{0.78 $\pm$ 0.02} & \textbf{0.86 $\pm$ 0.02} & 0.56 $\pm$ 0.04 \\
DINO v2 + MLP \textbar{} FT & \textbf{0.66 $\pm$ 0.03} & 0.44 $\pm$ 0.03 & 0.68 $\pm$ 0.02 & \textbf{0.78 $\pm$ 0.01} & \textbf{0.44 $\pm$ 0.05} & & \textbf{0.76 $\pm$ 0.02} & 0.58 $\pm$ 0.04 & 0.75 $\pm$ 0.01 & 0.83 $\pm$ 0.01 & 0.55 $\pm$ 0.03 & & \textbf{0.76 $\pm$ 0.02} & 0.58 $\pm$ 0.04 & \textbf{0.77 $\pm$ 0.03} & \textbf{0.84 $\pm$ 0.02} & \textbf{0.59 $\pm$ 0.05} \\
DINO v3 + MLP \textbar{} FT & 0.64 $\pm$ 0.02 & 0.44 $\pm$ 0.02 & 0.66 $\pm$ 0.03 & 0.76 $\pm$ 0.01 & 0.41 $\pm$ 0.05 & & 0.73 $\pm$ 0.03 & 0.57 $\pm$ 0.03 & 0.75 $\pm$ 0.02 & 0.83 $\pm$ 0.02 & \textbf{0.55 $\pm$ 0.04} & & 0.72 $\pm$ 0.02 & 0.57 $\pm$ 0.03 & 0.74 $\pm$ 0.02 & 0.82 $\pm$ 0.01 & 0.53 $\pm$ 0.02 \\
DINO v2 + MLP \textbar{} no FT & \textbf{0.68 $\pm$ 0.02} & 0.44 $\pm$ 0.03 & \textbf{0.70 $\pm$ 0.03} & \textbf{0.79 $\pm$ 0.02} & \textbf{0.48 $\pm$ 0.06} & & \textbf{0.77 $\pm$ 0.02} & 0.57 $\pm$ 0.03 & 0.76 $\pm$ 0.02 & 0.84 $\pm$ 0.01 & \textbf{0.54 $\pm$ 0.04} & & \textbf{0.76 $\pm$ 0.02} & 0.58 $\pm$ 0.03 & \textbf{0.76 $\pm$ 0.02} & \textbf{0.84 $\pm$ 0.01} & 0.55 $\pm$ 0.03 \\
DINO v3 + MLP \textbar{} no FT & 0.64 $\pm$ 0.02 & 0.44 $\pm$ 0.03 & 0.65 $\pm$ 0.02 & 0.75 $\pm$ 0.02 & 0.41 $\pm$ 0.05 & & 0.73 $\pm$ 0.03 & 0.57 $\pm$ 0.02 & \textbf{0.74 $\pm$ 0.02} & \textbf{0.83 $\pm$ 0.03} & \textbf{0.55 $\pm$ 0.04} & & 0.72 $\pm$ 0.01 & 0.57 $\pm$ 0.03 & 0.74 $\pm$ 0.01 & 0.82 $\pm$ 0.01 & 0.53 $\pm$ 0.02 \\

\bottomrule
\end{tabular}%
}
\end{table*}

\newcommand{\metricpanel}[2]{%
    \begin{subfigure}[t]{\linewidth}
        \centering
        \includegraphics[width=\linewidth]{#1}
        \caption{}
        \label{#2}
    \end{subfigure}%
}

\begin{figure*}[t]
    \centering
    \setlength{\tabcolsep}{3pt}
    \renewcommand{\arraystretch}{1}

    \begin{tabular}{
        @{}
        >{\centering\arraybackslash}m{0.07\textwidth}
        >{\centering\arraybackslash}m{0.30\textwidth}
        >{\centering\arraybackslash}m{0.30\textwidth}
        >{\centering\arraybackslash}m{0.30\textwidth}
        @{}
    }
        & \textbf{F1-score: Controls}
        & \textbf{F1-score: Patients}
        & \textbf{AUC}
        \\[0.5em]

        \textbf{MS}
        & \metricpanel
            {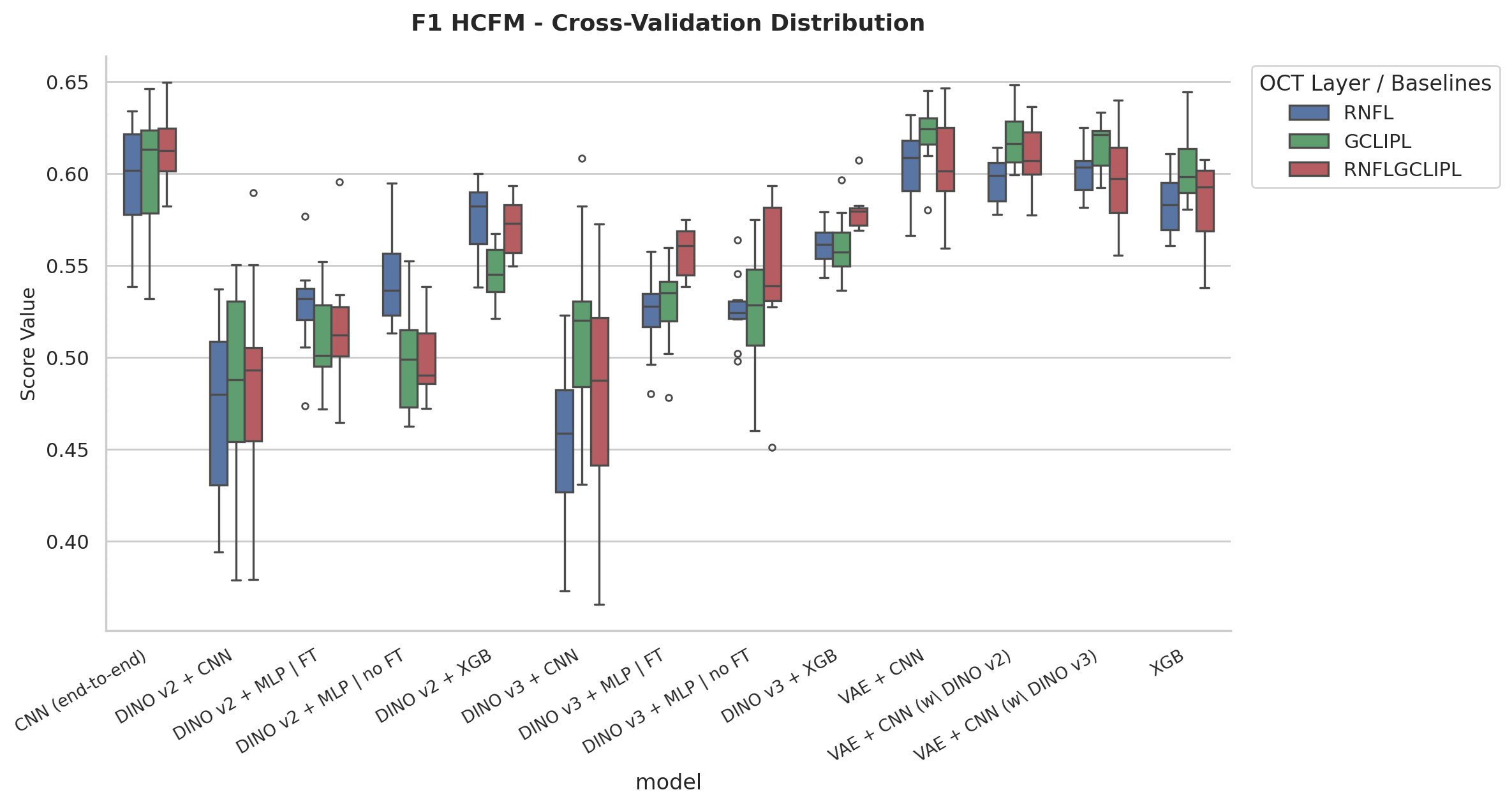}
            {fig:ms_hc_f1}
        & \metricpanel
            {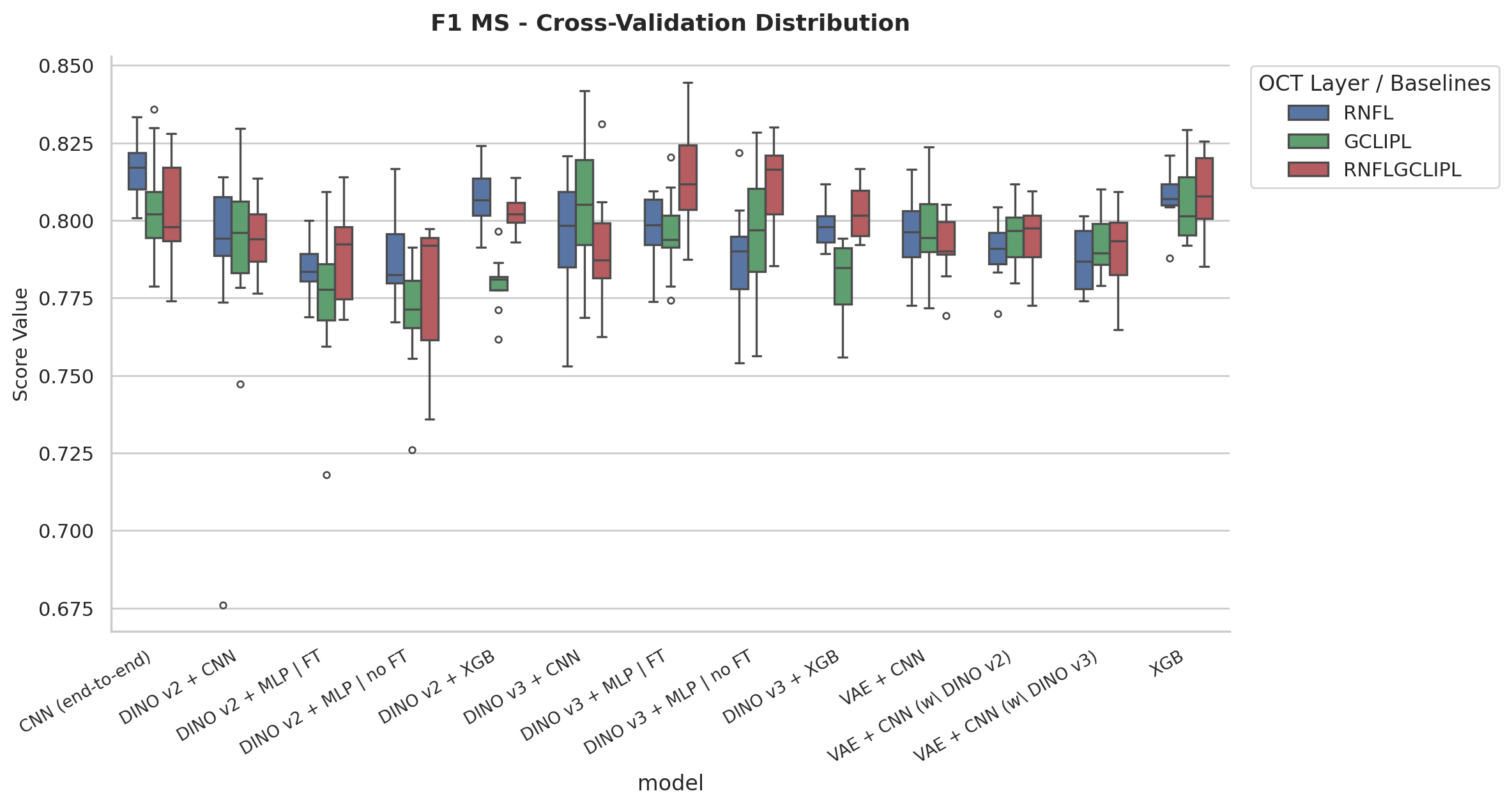}
            {fig:ms_patient_f1}
        & \metricpanel
            {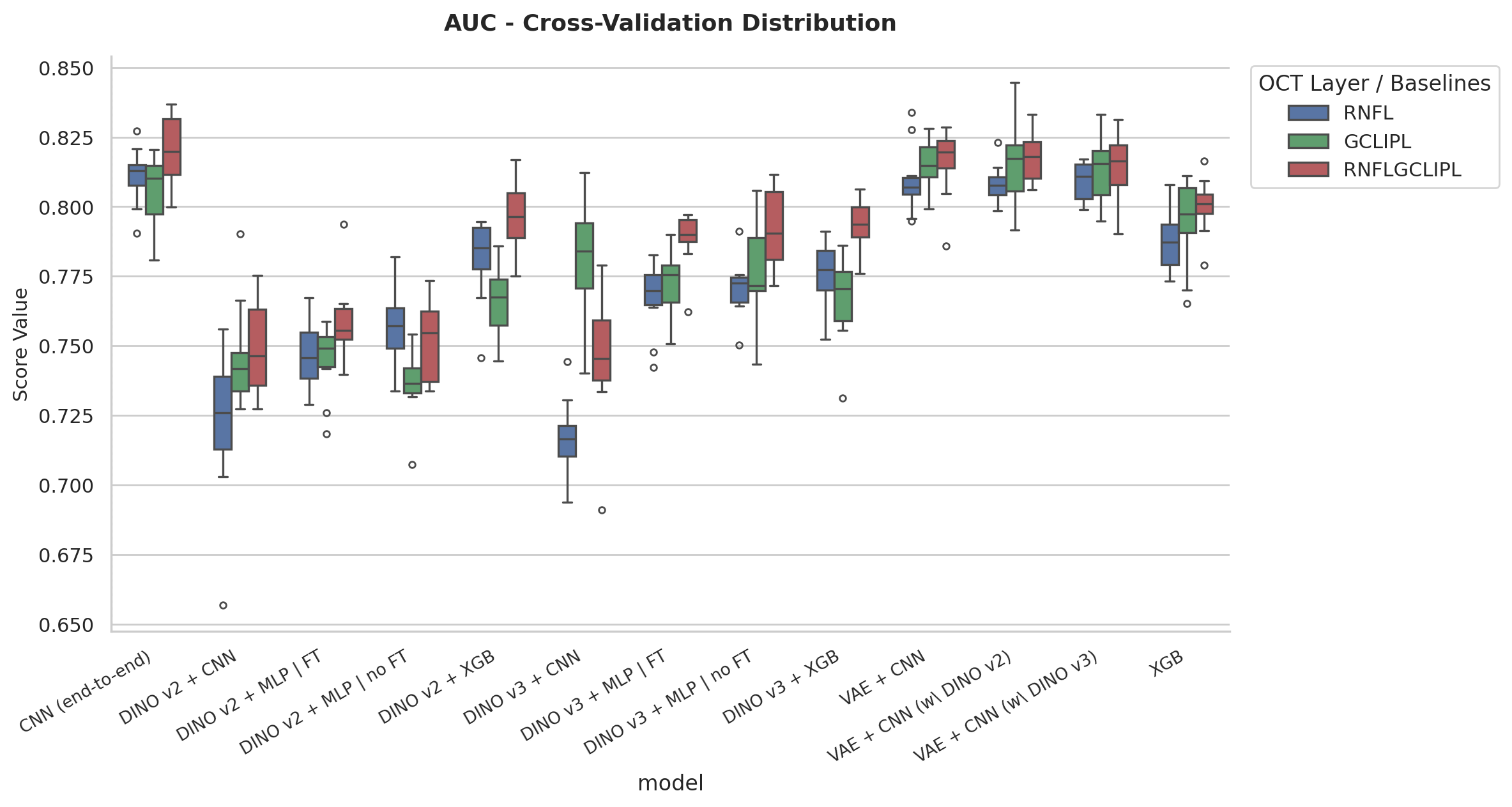}
            {fig:ms_auc}
        \\[0.8em]

        \textbf{PD}
        & \metricpanel
            {figs/example/f1_comparison/hcpd/boxplot_f1_HCfm.png}
            {fig:pd_hc_f1}
        & \metricpanel
            {figs/example/f1_comparison/hcpd/boxplot_f1_PD.png}
            {fig:pd_patient_f1}
        & \metricpanel
            {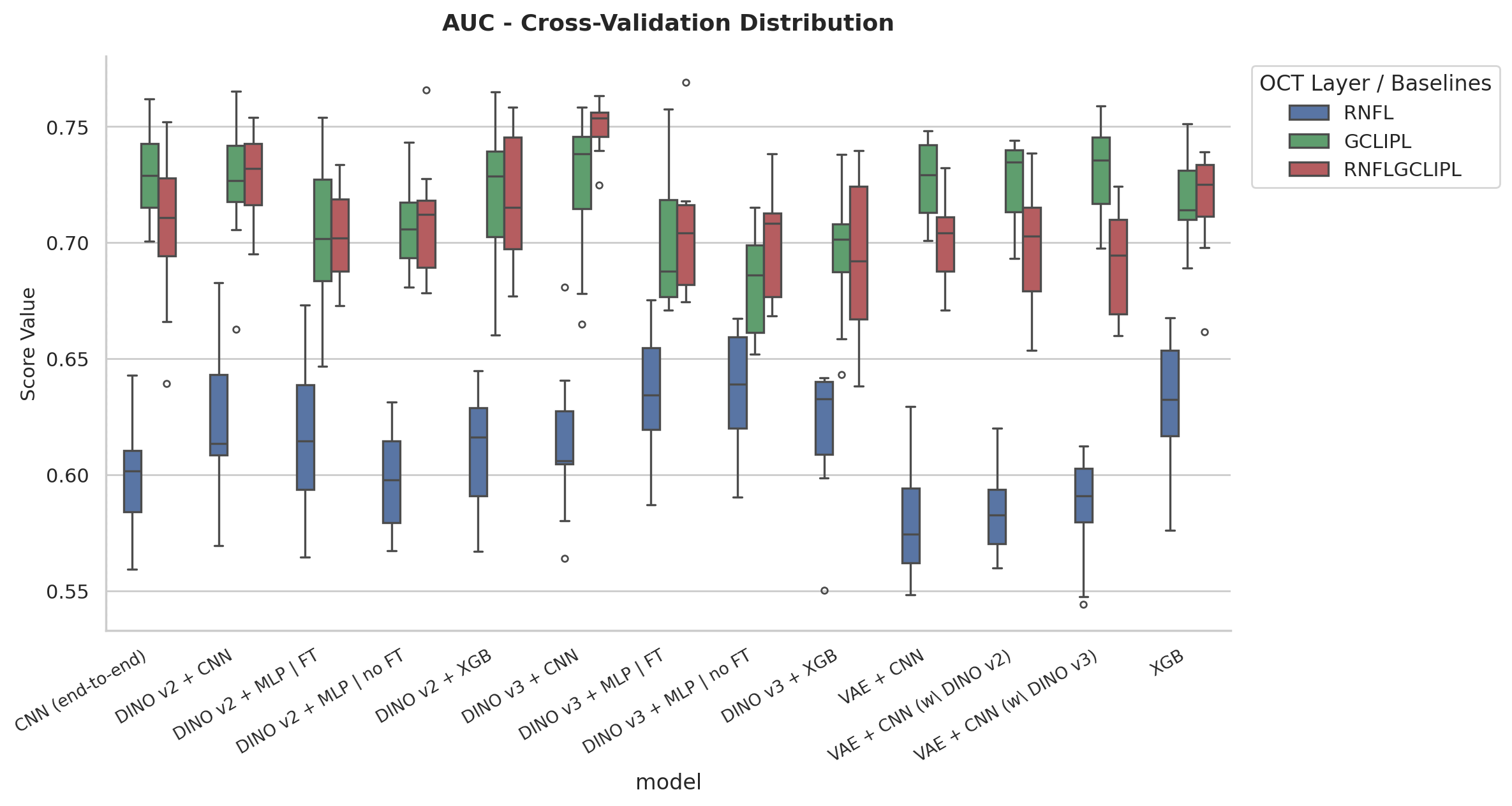}
            {fig:pd_auc}
        \\[0.8em]

        \textbf{AD}
        & \metricpanel
            {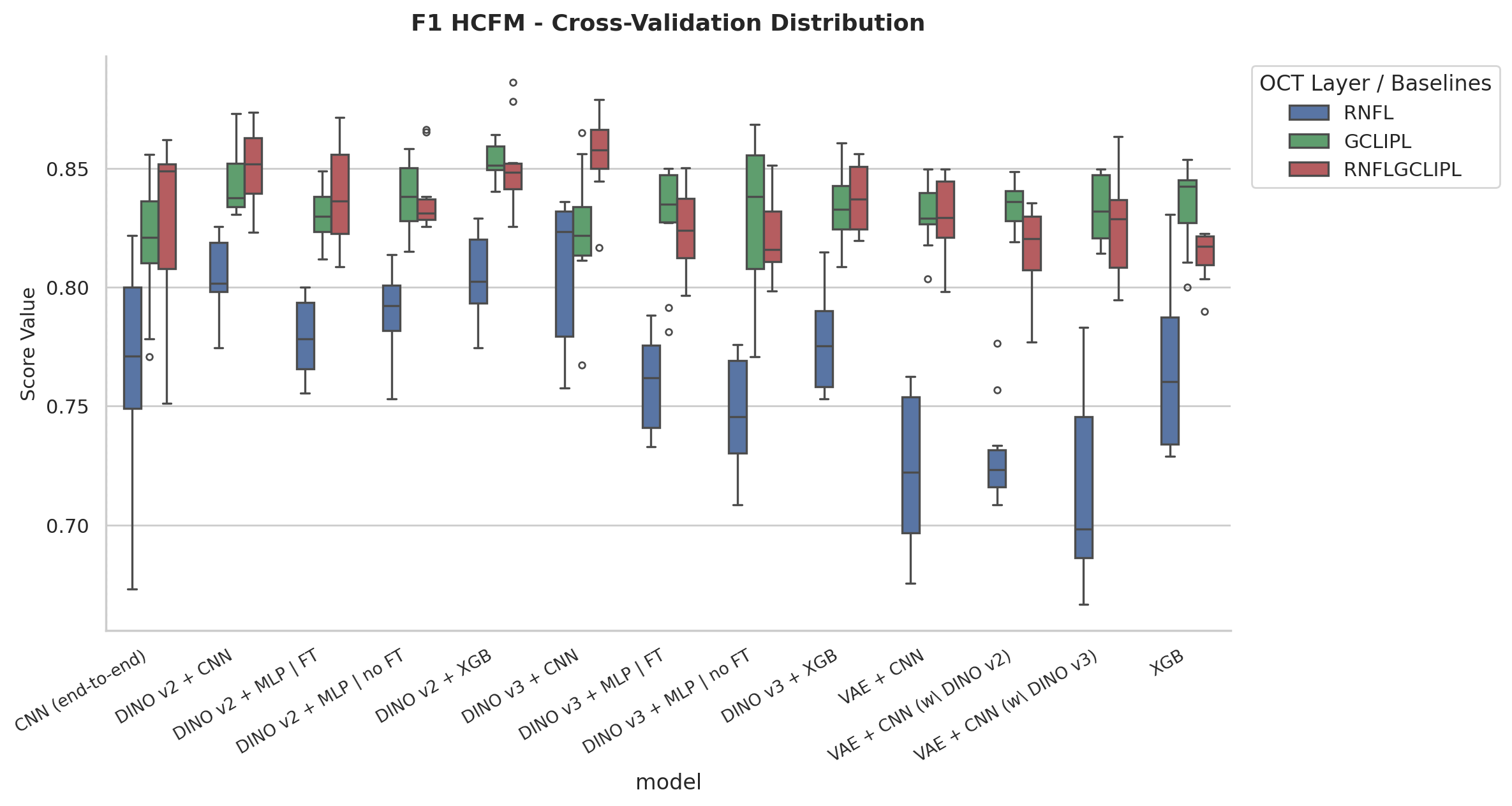}
            {fig:ad_hc_f1}
        & \metricpanel
            {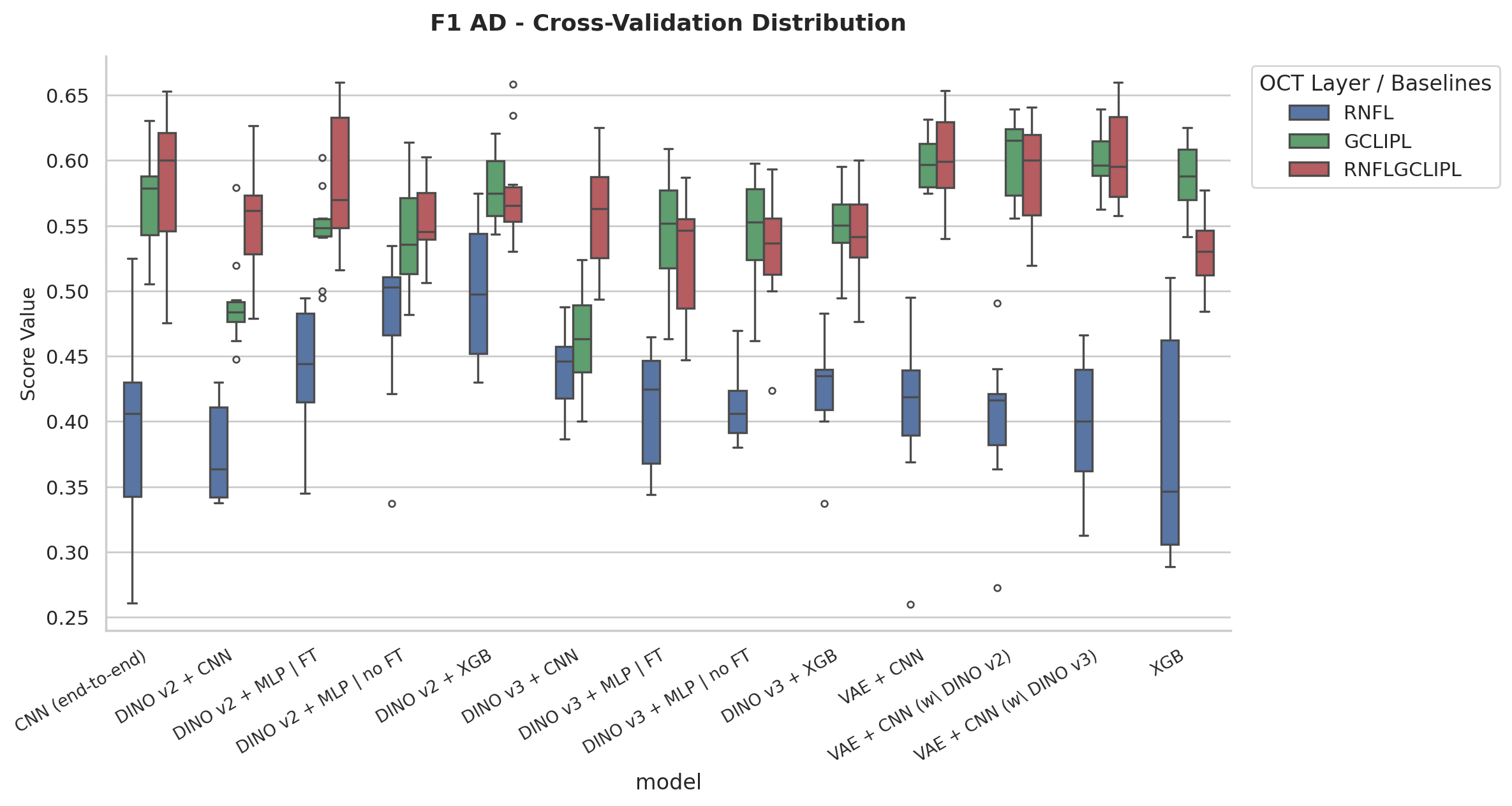}
            {fig:ad_patient_f1}
        & \metricpanel
            {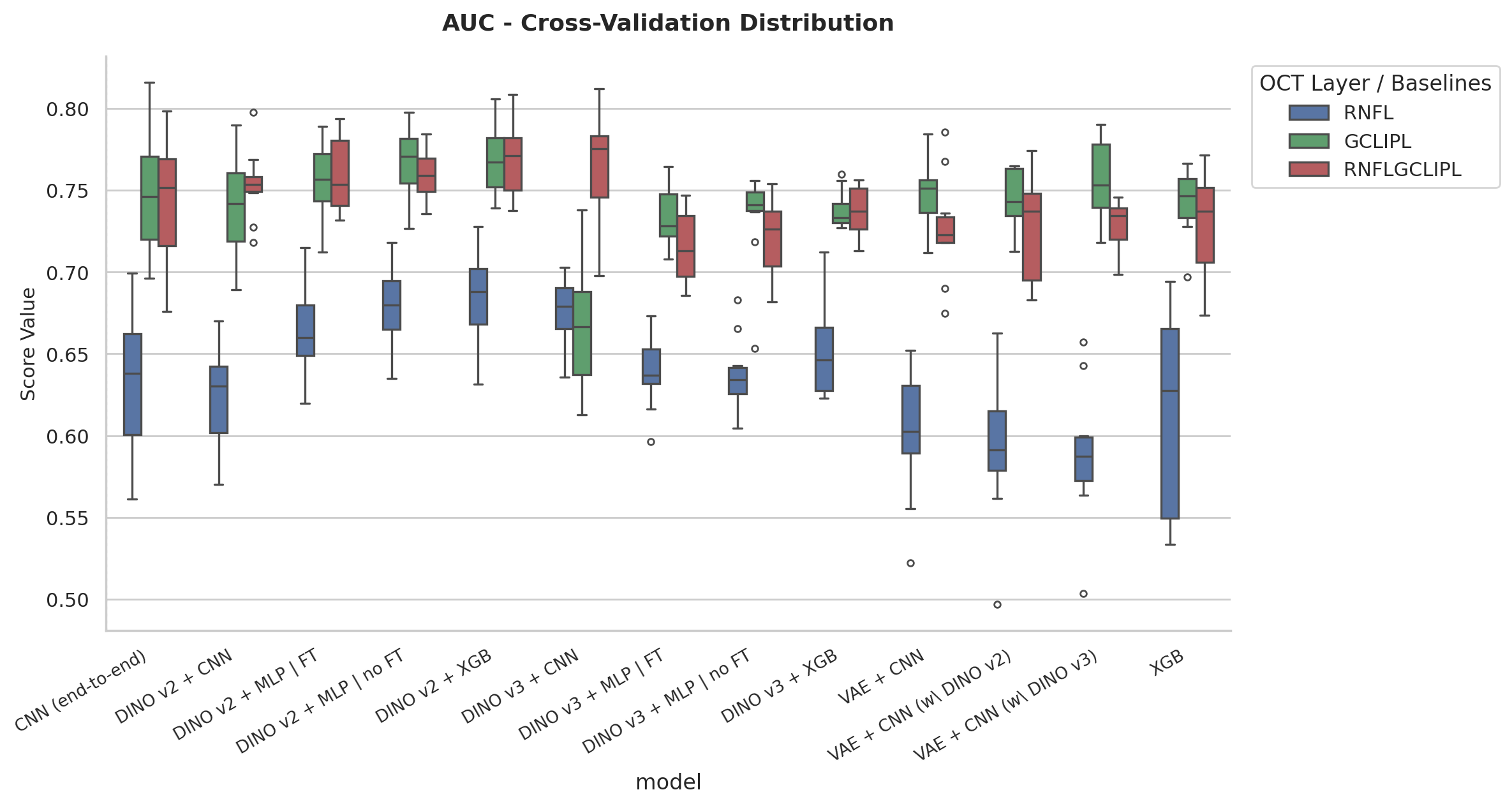}
            {fig:ad_auc}
        \\
    \end{tabular}

    \caption{Boxplots comparing the distribution of F1 scores for HC (First column) and patient classes (Second column), as well as AUC values (Third column) across all evaluated models for the three datasets: MS (Top row), PD (Middle row), and AD (Bottom row).}
    \label{fig:all_datasets_boxplots}
\end{figure*}

\subsection{FastMacula HC vs MS}

The first experiment was designed to evaluate the classification performance of healthy controls (HC) against multiple sclerosis (MS) patients using the FastMacula dataset from Miguel Servet Hospital (298 MS vs. 118 HC samples). Performance metrics are summarized in Top Table from~\ref{tab:comprehensive_all_datasets}. Overall, all evaluated classifiers achieved high and stable performance in the identification of MS patients, with $F_1$-MS values ranging between approximately 0.75 and 0.82. In contrast, we observed greater variability in the healthy control class, where $F_1$-HC scores varied substantially across architectures. Among the evaluated approaches, end-to-end CNN and VAE-based pipelines achieved the most balanced performance, reaching $F_1$-HC values around $0.60-0.62$. Fine-tuning of DINO representations produced only small differences compared with frozen feature extractors. For example, DINOv2 + MLP in the GCL-IPL configuration improved from $F_1$-HC = 0.50 to 0.51 after the parameter-efficient fine-tuning step. We found that no model was consistently superior with respect to the others, but rather that the discriminative information present in both RNFL and GCL-IPL layers was relevant for MS detection.

Regarding retinal layer configurations, both RNFL and GCL-IPL achieved competitive performance independently. However, GCL-IPL and RNFL + GCL-IPL configurations generally provided more stable results. The highest $F_1$-HC values were obtained by the VAE ($0.62 \pm 0.02$) and XGBoost ($0.60 \pm 0.02$) models when GCL-IPL information was included. As shown in Top row from Figure~\ref{fig:all_datasets_boxplots}, $F_1$-MS distributions remained consistently high across all methods, whereas $F_1$-HC exhibited larger variability, particularly for DINO-based architectures.

\subsection{FastMacula HC vs PD}

The second experiment evaluated the classification of healthy controls (HC) against Parkinson's disease (PD) patients (158 PD vs. 118 HC samples). Results are summarized in Middle Table from~\ref{tab:comprehensive_all_datasets}. Compared with the MS benchmark, the PD task proved to be more challenging, with the best-performing models achieving a maximum AUC of $0.75 \pm 0.01$. In this experiment, we clearly see a dependence on retinal layer selection: using standalone RNFL configuration resulted in a reduced performance across all architectures, with AUC values ranging approximately between 0.60 and 0.63. In contrast, the inclusion of GCL-IPL information increased performance across all evaluated models, giving AUC values up to $0.75 \pm 0.01$ and accuracies of $0.69 \pm 0.03$.

Among all evaluated configurations, DINOv3 + CNN achieved the highest overall performance in the RNFL + GCL-IPL setting, obtaining an AUC of $0.75 \pm 0.01$, AP of $0.79 \pm 0.01$, and Accuracy of $0.69 \pm 0.03$. This model was identified as the statistical leader according to the Holm post-hoc test ($p < 0.05$). The fine-tuning strategy again produced minimal performance differences: DINOv2 + MLP achieved an AUC of $0.70 \pm 0.03$ after fine-tuning and $0.71 \pm 0.02$ using frozen representations, and, under this experiment, the class-specific metrics showed relatively balanced performance between HC and PD categories. Middle row from Figure~\ref{fig:all_datasets_boxplots} illustrates that GCL-IPL-based models produced more compact and higher-performing distributions compared with the standalone RNFL configurations.

\subsection{FastMacula HC vs AD}

The third experiment evaluated the classification of healthy controls (HC) against Alzheimer's disease (AD) patients (46 AD vs. 118 HC). Results are presented in Bottom Table from~\ref{tab:comprehensive_all_datasets}. We can see a similar behavior as on the PD benchmark, where the GCL-IPL layer is more informative than the RNFL layer. For example, AUC values range from 0.59 to 0.69 in the RNFL configuration, while they increase to $0.67-0.77$ when GCL-IPL is included. The highest AUC values were obtained by DINOv2 + XGBoost and frozen DINOv2 + MLP within the GCL-IPL configuration ($0.77 \pm 0.02$). VAE-based approaches achieved the highest minority-class detection performance, reaching $F_1$ values of $0.60 \pm 0.02$ on the pathological class. Combining RNFL and GCL-IPL did not provide statistically significant gains over GCL-IPL alone, suggesting again the lower discriminative capabilities of the RNFL layer. However, some architectures, such as DINOv3 + CNN, benefited from the combined representation, improving AUC from $0.67 \pm 0.04$ to $0.76 \pm 0.03$. Similarly to the previous experiments, fine-tuning of DINO representations did not consistently improve performance, and DINOv3 did not consistently outperform its predecessor DINOv2. Bottom row from Figure~\ref{fig:all_datasets_boxplots} shows higher median AUC and $F_1$-AD values for GCL-IPL-based configurations compared with the standalone RNLF configuration models.

Importantly, given the constrained cohort size (46 AD subjects) and class imbalance, these findings should be interpreted as an exploratory proof-of-concept carrying less statistical weight than our MS and PD benchmarks. Beyond sample size, potential population biases must also be acknowledged: while AD predominantly affects older adults (60+ years), the broader age range in the control cohort introduces normal age-related retinal thinning as a confounding factor. Although class-weighted losses were applied during training, low AD sensitivity persists.

\subsection{ROC-AUC Curves Analysis}

%up and bottom collage
\begin{figure*}[t]
    \centering
    
    % Subfigura 1: MS
    \begin{subfigure}[b]{0.7\textwidth}
        \centering
        \includegraphics[width=\textwidth]{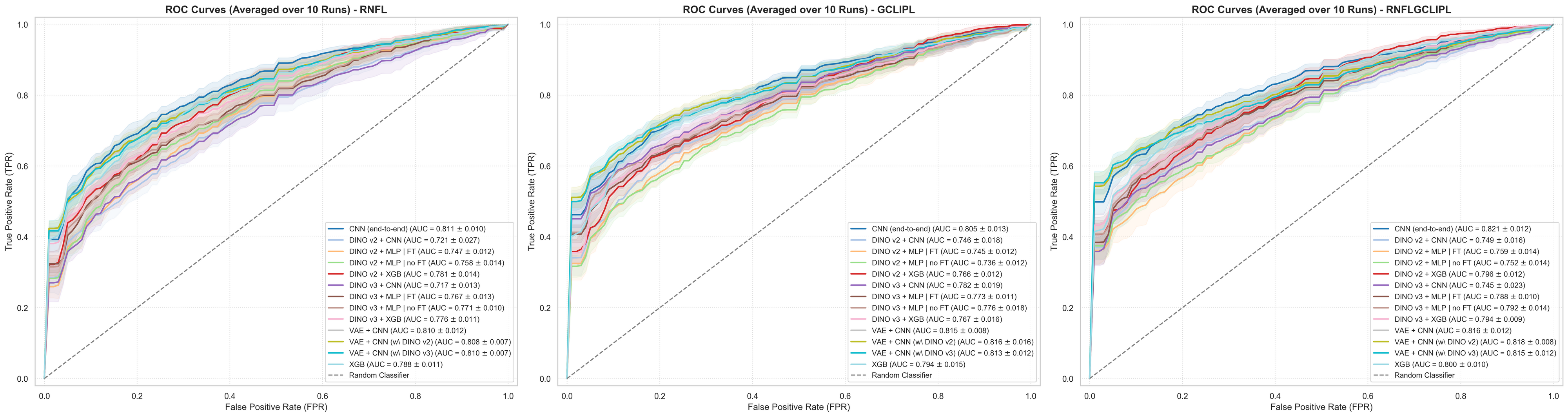} 
        \caption{Healthy Controls vs Multiple Sclerosis ROC curves across all runs.}
        \label{sub:hc_vs_ms_roc}
    \end{subfigure}
    
    \vspace{0.5cm}
    
    % Subfigura 2: PD (CORREGIDA: Ahora cierra correctamente)
    \begin{subfigure}[b]{0.7\textwidth}
        \centering
        \includegraphics[width=\textwidth]{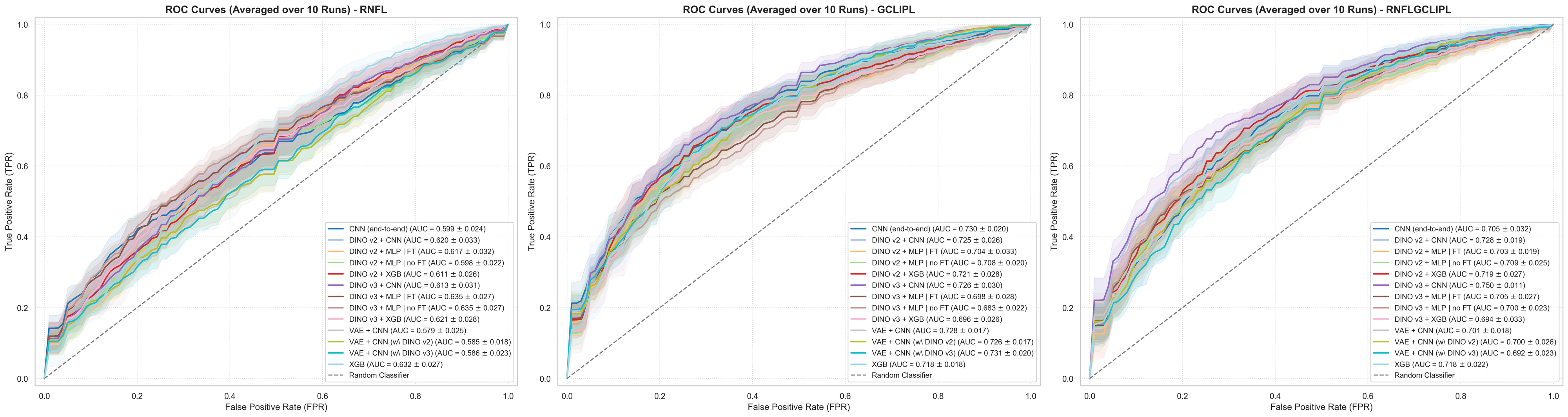}
        \caption{Healthy Controls vs Parkinson's Disease ROC curves across all runs.}
        \label{sub:hc_vs_pd_roc}
    \end{subfigure} % <-- Faltaba este cierre

    \vspace{0.5cm}
    
    % Subfigura 3: AD
    \begin{subfigure}[b]{0.7\textwidth}
        \centering
        \includegraphics[width=\textwidth]{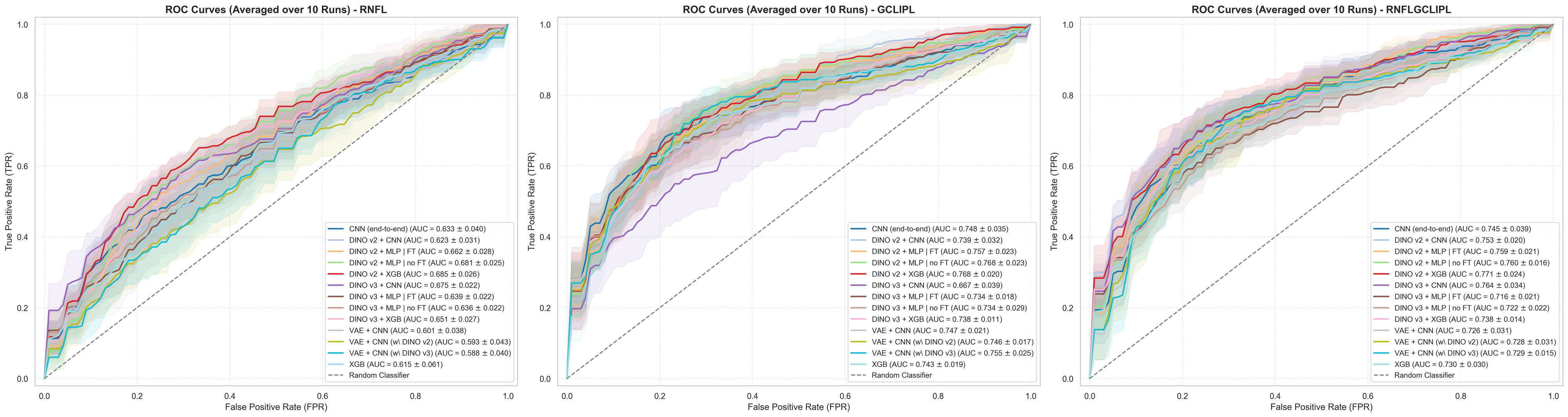}
        \caption{Healthy Controls vs Alzheimer's Disease ROC curves across all runs.}
        \label{sub:hc_vs_ad_roc}
    \end{subfigure}

    \caption{ROC curves for the classification of Healthy Controls (HC) from patients with Multiple Sclerosis (MS), Parkinson's Disease (PD), and Alzheimer's Disease (AD) for each input layer evaluated.}
    \label{fig:roc_curves_all}  
\end{figure*}

Cross-validated ROC curves in Figure~\ref{fig:roc_curves_all} clearly reflect these performance trends we have observed in the previous experiments. Across all layer configurations, classification performance for PD (b) and AD (c) remains lower than for MS (a). This confirms that while MS induces pronounced structural alterations across the retina, changes in PD and AD are subtler and more diffuse.

The ROC curves also mirror the layer dependencies previously identified. In subtle neurodegenerative scenarios (PD and AD), models trained only in RNFL underperform, whereas incorporating GCL-IPL (isolated or combined) substantially increases the performance curves. In contrast, for MS, both layers individually display solid discriminative trajectories, reinforcing that pathological damage affects both structures.

% Notably, a unique multi-layer effect occurs in the HC vs. AD task (Figure~\ref{fig:roc_curves_all}(c)). Under standalone GCL-IPL input, ROC curves display higher inter-model divergence and broader confidence intervals. Merging both layers (RNFL + GCL-IPL) eliminates this variance: curves cluster into a tighter, higher-performing bundle while confidence bands contract noticeably. While standalone GCL-IPL is sufficient in MS and PD, adding RNFL spatial context in the AD benchmark, which features a smaller, imbalanced cohort, acts as an effective implicit regularizer.

\subsection{Generalization and Explainability Analysis}

Beyond raw performance metrics, studying prediction uncertainty and model explainability is critical to verify whether deep architectures learn clinically meaningful retinal patterns rather than spurious artifacts. 

\subsubsection{Model Selection for the Analysis}

We centered this evaluation on a representative subset of architectures using the standalone GCL-IPL layer configuration, which consistently gave the most informative representations across all prior experiments. Furthermore, because DINO v3 did not give statistically significant performance gains over DINO v2, all DINO-based architectures were evaluated using the \textbf{DINO v2} backbone and also the fine-tuned version of the DINOv2 + MLP pipeline was not included in this analysis for the same reason. The selected architectures for this analysis were: 

\textit{(i)} Standalone XGBoost, 
\textit{(ii)} End-to-end CNN, 
\textit{(iii)} VAE + MLP (with DINOv2 Loss), 
\textit{(iv)} DINOv2 (Frozen) + MLP , 
\textit{(v)} DINOv2 (Frozen) + CNN, and 
\textit{(vi)} DINOv2 (Frozen) + XGBoost.

This generalization and explainability assessment is conducted across two complementary clinical scenarios:
\begin{enumerate}
    \item \textbf{HC vs. Multiple Sclerosis (MS):} An imbalanced scenario characterized by retinal neurodegeneration. Given the primary involvement of the inner retinal layers in MS, this task allows us to inspect whether models focus its attention on important retinal regions rather than on spurious artifacts.
    \item \textbf{HC vs. Parkinson's Disease (PD):} Serves as a subtle, early-stage pathology. Although PD-related retinal changes are clinically less evident, models trained on the GCL-IPL layer achieved remarkably strong classification metrics. Visualizing this more balanced cohort enables us to uncover the exact regions leveraged by the models to detect neurodegenerative signatures.
\end{enumerate}

To ensure a fair comparison, all models were evaluated on a fixed, stratified test split. For deep neural architectures with dropout layers (e.g., standard CNNs), we implemented a \textbf{Monte Carlo Dropout (MC Dropout)} strategy during inference by executing $T=50$ stochastic forward passes with active dropout. This allowed us to estimate prediction uncertainty (standard deviation), average classification metrics across passes, and compute consensus Grad-CAM++ maps. Architectures without dropout were evaluated using their native inference modes.

Explainability methods were adapted to the feature representations and gradient accessibility of each architecture. Standard end-to-end CNNs supported direct spatial attribution, whereas hybrid pipelines combining DINO backbones with MLPs or XGBoost required additional adaptations. These included spatial token reshaping and model wrapping for gradient-based explanations, and SHAP / TreeSHAP for feature-level attribution in non-differentiable classifiers.

%  Nota: esta seccion es necesaria? creo que no es muy informativa mas que para decir que generalizan todos robustamente?
\subsubsection{Quantitative Uncertainty Analysis}

\begin{table*}[t]
\centering
\caption{Quantitative performance metrics and prediction uncertainty evaluation across selected architectures (Only GCL-IPL layer).}
\label{tab:quantitative_mc_results}
\small
\setlength{\tabcolsep}{6pt} % Espaciado ideal para table*
\renewcommand{\arraystretch}{1.15}
\begin{tabular}{l cccc c cccc}
\toprule
& \multicolumn{4}{c}{\textbf{HC vs. MS}} & & \multicolumn{4}{c}{\textbf{HC vs. PD}} \\
\cmidrule(lr){2-5} \cmidrule(lr){7-10}
\textbf{Model Architecture} & \textbf{Acc. (\%)} & \textbf{$F_1$-HC} & \textbf{$F_1$-MS} & \textbf{$\sigma_{\text{MC}}$ (\%)} & & \textbf{Acc. (\%)} & \textbf{$F_1$-HC} & \textbf{$F_1$-PD} & \textbf{$\sigma_{\text{MC}}$ (\%)} \\
\midrule
Standalone XGB*             & 74 & 0.65 & 0.80 & ---  & & 78 & 0.74 & 0.82 & ---  \\
CNN (end-to-end)            & 70 & 0.61 & 0.75 & 3/9  & & 69 & 0.72 & 0.67 & 5/10 \\
VAE + MLP (w\textbackslash DINOv2) & 71 & 0.65 & 0.76 & 5/16 & & 74 & 0.67 & 0.78 & 8/15 \\
DINOv2 (Frozen) + CNN       & 75 & 0.68 & 0.78 & 7/11 & & 74 & 0.69 & 0.78 & 8/16 \\
DINOv2 (Frozen) + MLP       & 68 & 0.60 & 0.74 & 7/14 & & 67 & 0.56 & 0.73 & 9/17 \\
DINOv2 (Frozen) + XGB*      & 79 & 0.68 & 0.85 & ---  & & 76 & 0.72 & 0.79 & ---  \\
\bottomrule
\multicolumn{10}{l}{\footnotesize $\sigma_{\text{MC}}$: Monte Carlo Dropout prediction uncertainty expressed as mean / max standard deviation ($\mu_{\sigma} / \sigma_{\text{max}}$ in \%).} \\
\multicolumn{10}{l}{\footnotesize *Evaluated in standard deterministic mode (Monte Carlo Dropout non-applicable).}
\end{tabular}
\end{table*}

Table~\ref{tab:quantitative_mc_results} presents the quantitative test performance and Monte Carlo uncertainty metrics. Hybrid pipelines incorporating foundation features achieved strong diagnostic accuracy, led by DINOv2 (Frozen) + XGB in HC vs. MS ($79\%$ Acc, $F_1$-MS = $0.85$) and Standalone XGB in HC vs. PD ($78\%$ Acc). Conversely, frozen backbones combined with MLPs (DINOv2 (Frozen) + MLP) exhibited slight drops across tasks. 

Importantly, Monete Carlo Dropout uncertainty ($\sigma_{\text{MC}}$) remained consistently low across deep models, with mean standard deviations bounded below $\mu_{\sigma} \leq 9\%$. Maximum uncertainty ($\sigma_{\text{max}}$) was slightly higher in PD (up to $17\%$) than in MS ($16\%$), matching the subtle and diffuse nature of the Parkinson neurodegeneration. These low uncertainty bounds confirm that the models give confident, stable predictions, supporting their generalization capability on unseen test data. Notably, tree-based models (XGBoost) do not support Monte Carlo Dropout, but their deterministic predictions were stable and aligned with the deep architectures performance.

\subsubsection{Qualitative Visual Interpretation of Attribution Maps}

\begin{figure*}[t]
    \centering
    \includegraphics[width=\textwidth]{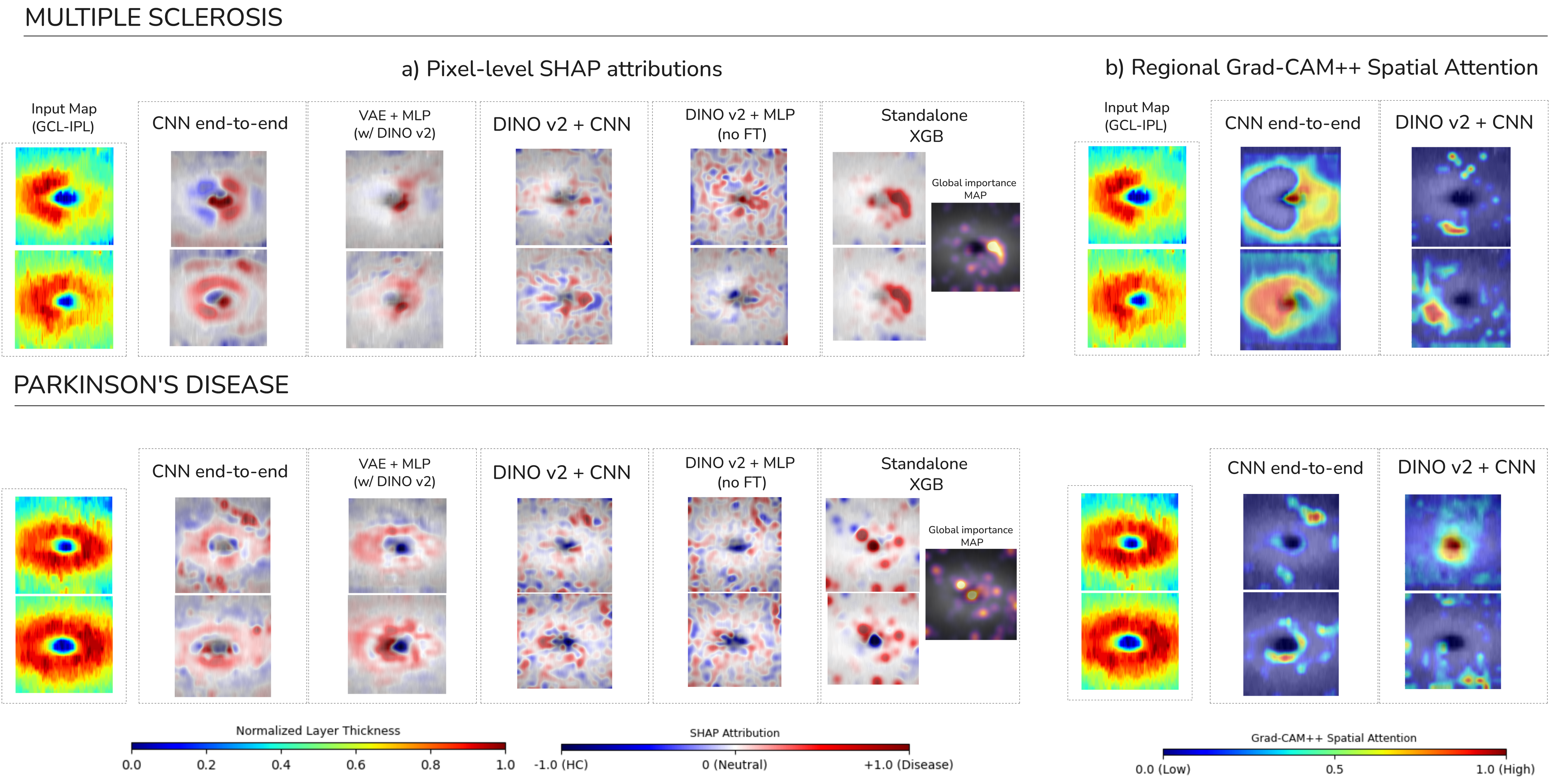}
    \caption{Comparative visual attribution maps for representative MS and PD patients across evaluated architectures, splitted into two interpretability strategies: \textbf{(a) Pixel-level SHAP attributions} evaluating end-to-end feature importance across all five model baselines, normalized to $[-1, 1]$ (blue: favors HC, red: favors pathology). \textbf{(b) Regional Grad-CAM++ Spatial Attention} capturing high-level activation regions across architectures featuring trainable convolutional heads (Standard CNN and DINOv2 + CNN), normalized to $[0, 1]$ (red: high attention). Raw GCL-IPL thickness maps are provided adjacent to each panel for spatial reference.}
    \label{fig:explainability_maps}
\end{figure*}

Figure~\ref{fig:explainability_maps} displays representative explainability maps for MS and PD patients. SHAP provides fine-grained directional feature importance (red: favors pathology, blue: favors control), whereas Grad-CAM++ delineates broad regional spatial focus.

Comparing architectures reveals structural differences in feature utilization:
\begin{itemize}
\item \textbf{DINO-based models vs. Image-based CNNs/VAEs:} DINO-based architectures display more diffuse SHAP attributions due to the Vision Transformer's patch-based tokens ($14 \times 14$) and non-local self-attention. Nevertheless, they consistently concentrate positive attributions (red) around the macular GCL-IPL region. In contrast, standard CNNs and VAEs generated highly localized, sharp regions of interest centered on macular thinning.
\item \textbf{Effect of Classification Heads:} DINO v2 + MLP produces spatially dispersed SHAP maps because the downstream MLP is working with 1D global features, whereas DINO v2 + CNN retains spatial correspondence, producing more focused Grad-CAM++ maps. On the other hand, the XGBoost classifier produces a highly sparse set of SHAP attributions due to its tree-based, threshold-driven feature selection, where non-selected pixels receive exact zero attributions.
\end{itemize}

From a clinical standpoint, attribution patterns align with pathological signatures: MS induces well-defined, focal attributions over the central macular region, explaining the superior performance of image-based models in this task. Conversely, PD exhibits a broader, more diffuse macular footprint. Finally, while Grad-CAM++ occasionally exhibits edge-activation artifacts on image boundaries, SHAP maps demonstrate greater robustness against boundary noise due to their direct pixel-to-output evaluation.

\section{Discussion}
\label{sec:discussion}

The automated detection and differential diagnosis of neurodegenerative conditions through non-invasive optical coherence tomography (OCT) macular en-face images remain a critical frontier in neuro-ophthalmology. In our previous work~\cite{morancavero2026spatialnormalizationcrossdomainretinal}, we obtained normalized retinal segmentations which allowed us now to build high resolution thickness maps. From these results, we evaluate different machine learning and deep learning approaches for the classification of three neurodegenerative diseases. These include traditional machine learning models, CNNs, generative models, and self-supervised Vision Transformers. Our results show promising classification performance across the studied diseases. Beyond numerical metrics, the explainability analysis allows us to visualize which retinal regions contribute most to the model predictions.

\subsection{Pathology-Specific Structural Biomarkers Across Retinal Layers}

A primary finding of this study is the decisive role of retinal layer selection in all evaluated tasks. In each pathological condition, models trained exclusively on the Retinal Nerve Fiber Layer (RNFL) got a severe performance degradation, characterized by collapsed AUC metrics and lower $F_1$-scores. Conversely, incorporating the ganglion cell complex (GCL-IPL or the combination of RNFL + GCL-IPL) consistently restored performance and stabilized variance across runs. This behavior correlates directly with clinical pathophysiology across conditions:

\begin{itemize}
    \item \textbf{Multiple Sclerosis (MS):} MS pathology is characterized by pronounced neurostructural damage. On this experiment, models achieved high and stable performance, with the best configurations reaching a peak AUC above $0.80$ using the GCL-IPL layer.
    \item \textbf{Parkinson's Disease (PD):} PD presents a more subtle and diffuse neurodegenerative footprint, resulting in a lower overall performance ceiling (peak AUC of $0.75 \pm 0.01$). Layer sensitivity was critical: RNFL models failed to exceed an AUC of $0.63$, whereas configurations incorporating the GCL-IPL layer elevated the performance to the peak AUC of $0.75$.
    \item \textbf{Alzheimer's Disease (AD):} In the highly imbalanced AD benchmark, the standalone RNFL configuration similarly suffered a collapse. In contrast, incorporating the GCL-IPL layer substantially restored performance, mirroring the robust improvements observed in the PD experiments, with foundation-based hybrid models yielding the top AUC metrics.
\end{itemize}

Interestingly, combining RNFL and GCL-IPL did not provide statistically significant improvements over standalone GCL-IPL configurations. This indicates that the majority of the discriminative information is already captured within the ganglion cell complex, while the RNFL primarily contributes redundant or marginal contextual information.

\subsection{Foundation Models, Architectural Complexity and Fine-Tuning}

An interesting finding across all experiments is that foundation models did not outperform simpler architectures. While DINO-based pipelines occasionally achieved the highest overall metrics, particularly in the PD benchmark, these improvements were not consistent across the other experiments, layer input, or evaluation metrics. In the MS cohort, conventional end-to-end CNNs and VAE-based representations achieved performance comparable to, or better than, DINO-derived features, particularly when considering minority-class performance (with class imbalance present through all the experiments). Nearly all architectures obtained statistically equivalent $F_1$-MS scores, while CNN and VAE pipelines maintained the highest $F_1$-HC values. This suggests that the discriminative information required for MS detection is already represented within retinal thickness maps, reducing the architectural complexity required for an effective classification pipeline. In contrast, the behavior observed in PD and AD was more subtle. In these pathologies, retinal alterations are less pronounced, making discrimination more challenging. Under these conditions, DINO-derived representations achieved statistical equivalence with traditional approaches and occasionally reached the top-performing configurations. Nevertheless, these gains were not significant enough and did not establish a clear superiority of foundation models over simpler architectures. Notably, the VAE-based representations consistently provided stable performance on the minority class across all the experiments.

These findings highlight an important consideration for clinical machine learning applications. Recent advances in foundation models have generated the expectation that increasing pretraining data and model capacity will automatically translate into superior downstream performance. However, our results suggest that this assumption does not necessarily hold in highly specialized medical imaging scenarios characterized by limited annotated datasets, domain-specific biomarkers, and class imbalance. In such settings, architectural simplicity and high-quality anatomically normalized inputs may be as important as model scale.

To further investigate the role of backbone adaptation, we evaluated Low-Rank Adaptation (LoRA) as a parameter-efficient fine-tuning strategy for DINOv2. By preserving the pretrained backbone weights and optimizing only a small number of additional trainable parameters, LoRA enables task-specific adaptation while limiting the number of parameters updated during training. However, this strategy did not produce statistically significant improvements over the frozen backbone configurations. One possible interpretation is that the pretrained representations already capture features relevant to the downstream classification tasks, leaving limited improvement margin through parameter-efficient adaptation under the available training conditions. Alternatively, the limited cohort sizes and subtle retinal alterations associated with certain pathologies may constrain the benefits of fine-tuning. 

\subsection{Explainability and interpretability}

Explainability analyses provided supporting evidence that the models focused on anatomically meaningful retinal regions. The use of high-resolution en-face thickness maps derived from retinal layer segmentations enabled spatially detailed attribution analyses, allowing us to analyze localized patterns of model relevance that could be obscured by conventional regionally averaged thickness measurements. Additionally, Monte Carlo Dropout ($T = 50$) produced low average predictive variability ($\mu_\sigma \leq 9\%$) across the evaluated deep learning models, suggesting relatively stable predictions under stochastic inference. Interestingly, the attribution maps revealed different spatial patterns between pathologies: MS models exhibited more localized regions of relevance within the macula, whereas PD models showed a more diffuse distribution across the macular area.

These differences are consistent with patterns described in the clinical literature on retinal neurodegeneration. In multiple sclerosis (MS), neuroretinal thinning can exhibit region-specific patterns of axonal and neuronal damage. Zahavi et al.~\cite{zahavi2025macular} reported regional differences in inner retinal alterations, including specific macular sectors within the GCL-IPL. Conversely, Parkinson's disease (PD) has been associated with more spatially distributed alterations affecting inner retinal structures, including the GCL-IPL and INL~\cite{wagner2023retinal}. Although attribution maps do not directly establish the presence or anatomical distribution of retinal damage, the observed spatial differences suggest that the high-resolution en-face representations may facilitate the exploration of disease-associated morphological patterns beyond predefined regional measurements.

\subsection{Limitations and Future Work}

Despite the robust statistical framework and comprehensive multi-model evaluation, several limitations of this study must be acknowledged. First, the evaluation relies on a single-center cohort with notable sample size constraints and class imbalances, most prominently in the Alzheimer's Disease cohort. This limited dataset scale introduces a potential risk of downstream classifier overfitting, particularly when training on high-dimensional feature spaces extracted from large foundation backbones. Consequently, validation across multiple domains, diverse acquisition protocols, and different OCT acquisition devices remains essential to confirm cross-site generalization. Second, generating pixel-wise SHAP attributions across stochastic Monte Carlo Dropout runs ($T = 50$) introduces a substantial computational overhead that could restrict real-time integration into clinical workflows. Finally, the non-differentiable nature of tree-based classifiers (XGBoost) requires feature flattening or standalone TreeSHAP evaluations, preventing the direct generation of smooth, continuous 2D attribution maps when paired with frozen Vision Transformer backbones.

To address these constraints and build upon our findings, future research will naturally extend along several broader methodological axes. Rather than treating en-face thickness maps in isolation, subsequent studies could investigate more data modalities or alternative scanning protocols to further test representational limits. Furthermore, expanding validation to larger, multi-center cohorts will be essential to establish broader clinical generalizability and mitigate the effects of cohort-specific biases identified in exploratory analyses like the Alzheimer's Disease benchmark. Finally, exploring alternative self-supervised adaptation strategies will remain an ongoing effort as newer foundation models emerge in medical imaging.

\section{Conclusion}
\label{sec:conclusion}

In this work, we presented a benchmark evaluating traditional machine learning classifiers, deep neural networks, VAE-based pipelines, self-supervised foundation backbones, and explainability frameworks for the classification of healthy controls from a subset of neurodegenerative diseases individually. Using high-resolution 2D macular en-face thickness maps derived from anatomically normalized OCT segmentations, our results align with other clinical findings highlighting the relevance of the GCL-IPL complex as a discriminative retinal biomarker while the configurations with only the RNFL layer generally showed lower predictive performance.

Among the evaluated approaches, conventional CNNs and VAE-based pipelines achieved competitive performance compared with large-scale self-supervised models, despite their lower architectural complexity, highlighting that more complex architectures do not necessarily lead to better classification performance in small and imbalanced clinical datasets. Furthermore, the use of high-resolution en-face thickness maps enabled spatially detailed interpretability analyses, allowing us to investigate whether model predictions were associated with anatomically meaningful retinal regions.

\section*{Acknowledgments}
This work was partially supported by the national research grants PID2022-138703OB-I00 (Trust-BEyE project) and PID2023-148219OB-C22, funded by MCIN/AEI/10.13039/501100011033/FEDER, UE; RICORS network of inflammatory diseases RD24/0007/0022 from Carlos III Health Institute; Government of Aragon grant PROY B50 24 and Group Reference T64 26R (COS2MOS research group); European HORIZON-MSCA-2024-SE-01 G.A. 10123661; and European Chistera action CLASiK - PCI2025-167187-2. The funders had no role in study design, data collection and analysis, decision to publish, or preparation of the manuscript.

\bibliographystyle{elsarticle-num}
\bibliography{references}

\end{document}